\documentclass[10pt,journal]{IEEEtran}

\usepackage[T1]{fontenc}
\usepackage[utf8]{inputenc}
\usepackage{cite}
\usepackage{amsmath}
\usepackage{amssymb}
\usepackage{microtype}
\usepackage{url}
\usepackage{booktabs}
\usepackage{tabularx}
\usepackage{graphicx}
\usepackage{xcolor}
\usepackage{balance}

\newcolumntype{L}[1]{>{\raggedright\arraybackslash}p{#1}}
\newcolumntype{Y}{>{\raggedright\arraybackslash}X}

\title{{\scshape Channels4World}: An Environment-Scale Dataset for Wireless Foundation Models}
\author{Joonkyu Jang,~\IEEEmembership{Student Member,~IEEE},
Hyung-Joo Moon,~\IEEEmembership{Member,~IEEE}, Kwang Soon Kim,~\IEEEmembership{Senior Member,~IEEE}, Seong-Lyun Kim,~\IEEEmembership{Senior Member,~IEEE}, Jakob Hoydis,~\IEEEmembership{Fellow,~IEEE},\\ Robert W. Heath, Jr.,~\IEEEmembership{Fellow,~IEEE}, and
Chan-Byoung Chae,~\IEEEmembership{Fellow,~IEEE}
\\\vspace{1.5em} \textit{Invited Paper}
\thanks{J. Jang, H.-J. Moon, and C.-B. Chae are with the School of Integrated Technology, Yonsei University, Seoul 03722, South Korea (e-mail: \{joonkyuj, moonhj, cbchae\}@yonsei.ac.kr).}
\thanks{K. S. Kim and S.-L. Kim are with the School of Electrical and Electronic Engineering, Yonsei University, Seoul 03722, South Korea (e-mail: \{ks.kim, slkim\}@yonsei.ac.kr).}
\thanks{J. Hoydis is with NVIDIA, 92400 Courbevoie, France (e-mail: jhoydis@nvidia.com).}
\thanks{R. W. Heath, Jr. is with the Department of Electrical and Computer Engineering, University of California San Diego, La Jolla, CA 92093 USA (e-mail: rwheathjr@ucsd.edu).\\
\emph{Corresponding author: Chan-Byoung Chae}.}
}

\IEEEpubid{0000-0000/00\$00.00~\copyright~2026 IEEE}

\begin{document}
\maketitle

\begin{abstract}
Wireless foundation models aim to learn reusable representations that transfer across tasks, configurations, and deployment environments.
Their pretraining depends on environmental breadth as well as observation count, yet existing wireless datasets often provide dense observations from few sites.
In this paper, we introduce {\sffamily\scshape Channels4World}, a scene-centered dataset of 100,000 distinct outdoor radio-frequency (RF) scenes and a procedural methodology for generating them.
Each scene binds one procedurally generated environment and its RF material assignments to 32 base-station (BS) positions, 16,384 user-equipment (UE) positions, and aligned variable-length multipath and ordered interaction records at 3.5 and 7~GHz.
Across the corpus, 52.43 billion physical BS--UE pairs produce 104.86 billion frequency-specific evaluations in a 29.14-TB packaged payload.
Scene identity keeps geometry, placements, channels, and interactions aligned across frequencies while allowing environmental breadth and within-environment observation density to be varied separately.
The public release provides predefined subsets at several scales and separate access to corpus components.
A deterministic 90K/10K scene-level train--test split keeps both frequencies and all components of each scene together.
We characterize morphology, RF material composition, and transceiver placement across three generation profiles.
Controlled propagation and path-retention analyses show that diffraction materially changes retained-link availability and multipath structure, while the corpus retention policy substantially reduces storage with little loss in represented aggregate power.
Under matched training exposure, models trained on broader scene collections achieve better held-out-scene performance, although an equal-episode control indicates that this gain cannot be attributed to scene count alone.
By making environmental breadth explicit and controllable, {\sffamily\scshape Channels4World} provides a large-scale resource for developing and evaluating wireless foundation models intended to generalize across deployment environments.
\end{abstract}


\IEEEpeerreviewmaketitle

\newpage
\section{Introduction}

\IEEEPARstart{W}{ireless} propagation is strongly conditioned by the physical environment.
Its geometry shapes blockage and path visibility, while its material properties affect the electromagnetic interactions between radio waves and surrounding structures.
Within a given environment, base-station (BS) and user-equipment (UE) positions determine which propagation paths contribute to each link.
Varying these positions can produce many distinct channel observations without adding new environments.
A large aggregate link count therefore does not by itself imply broad environmental coverage.
We distinguish two complementary dimensions of wireless-data scale: within-environment observation density and environmental breadth.
Dense sampling resolves local spatial variation, whereas sampling across environments exposes geometric and material variation needed to study transfer beyond familiar sites.

\begin{figure*}[t]
    \centering
    \includegraphics[width=\textwidth]{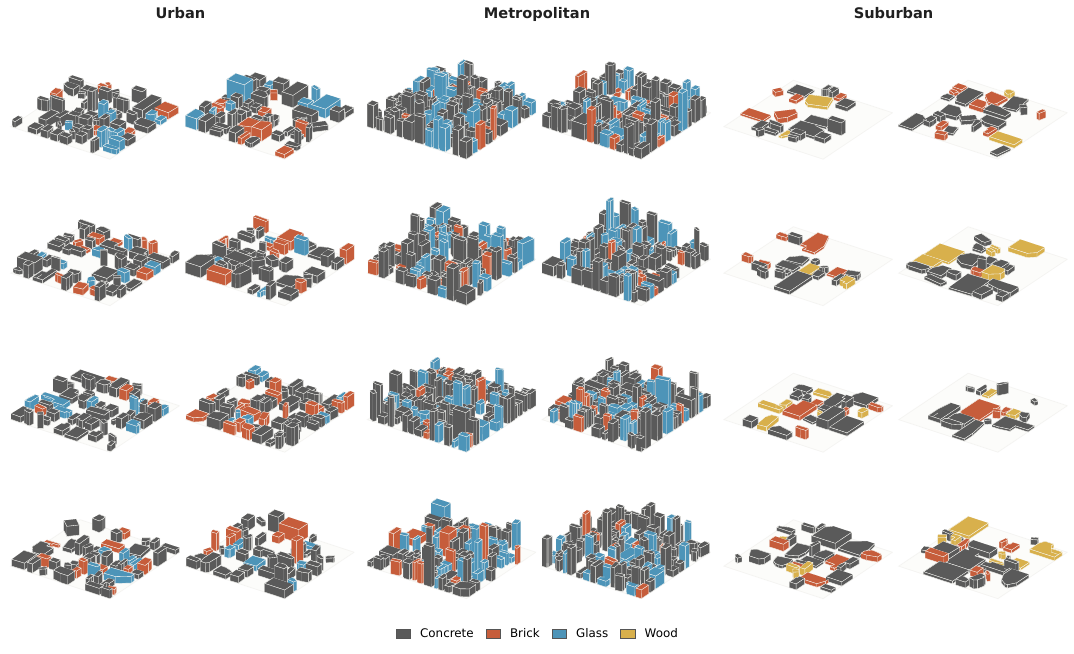}
    \caption{Example {\sffamily\scshape Channels4World} environments across the Urban, Metropolitan, and Suburban generation profiles.
    Building colors denote assigned RF material classes.}
    \label{fig:channels4world_at_a_glance}
\end{figure*}

\IEEEpubidadjcol

This distinction becomes increasingly consequential as artificial intelligence (AI) assumes a larger role in emerging 6G systems~\cite{itu_imt2030}.
Wireless foundation models aim to learn reusable representations that transfer across wireless tasks and deployment settings rather than specialize to a single task or site~\cite{jiang2025cfm}.
The diversity of the pretraining data therefore matters, not just its volume.
A model trained on millions of observations from only a few environments may learn detailed site-specific regularities without learning how propagation structure changes across environments.
Pretraining intended to support reuse across environments therefore benefits from large observation sets spanning varied environments and from evaluation protocols that distinguish interpolation within familiar environments from generalization to unseen ones.
Wireless foundation models provide the primary motivating case here, but the same breadth--density distinction applies more broadly to data-driven methods intended to transfer across deployment environments.

Existing wireless-data resources address different parts of this objective.
Configurable statistical and geometry-based channel models support controlled experiments.
Site-specific ray tracing provides dense channels tied to explicit geometry, while radio-map and multimodal datasets preserve spatial or environmental context~\cite{alkhateeb2019deepmimo,klautau2021raymobtime,huangfu2022waird,levie2021radiounet}.
Real-world measurements capture hardware effects, interference, dynamics, and propagation behavior that synthetic models may not reproduce~\cite{alkhateeb2023deepsense}.
However, scaling these approaches to many self-contained environments under common conventions remains difficult.
Simulation requires suitable environment assets, deployment definitions, material assignments, and validation, while measurement campaigns require substantial equipment, calibration, access, and post-processing.
Combining independently produced resources can also introduce inconsistencies in channel representations, thresholds, coordinate systems, and environmental metadata~\cite{shakya2025pointdata}.

An environment-scale wireless corpus must combine broad environmental coverage with dense observations within each environment in a consistent representation that preserves alignment among the physical environment, transceiver deployment, and resulting channel records.
To address these requirements, we introduce {\sffamily\scshape Channels4World}, a scene-centered dataset and procedural generation methodology built around self-contained outdoor scene records.
The procedural pipeline constructs environment geometry, assigns radio-frequency (RF) materials, samples environment-informed BS/UE placements, and produces the corresponding channel and interaction records.
This organization makes environment count and within-environment observation density independently controllable.

The corpus comprises 100,000 scenes: 50,000 Urban, 25,000 Metropolitan, and 25,000 Suburban---these labels denote configured generation profiles rather than universal classes of real cities.
Each scene binds one procedurally generated environment and its RF material assignments to 32 BS positions and 16,384 UE positions, with the same placement set evaluated at 3.5 and 7~GHz.
Across the corpus, this yields 52.43 billion physical BS--UE pairs and 104.86 billion frequency-specific evaluations.
GPU-accelerated ray tracing with Sionna RT produces the multipath channel data.
A fixed path-retention policy keeps at most the 16 strongest candidate paths that also exceed a power threshold, yielding compact, variable-length multipath records.
These components share a common scene identity, allowing geometry, placements, frequency-specific channel records, and ordered interaction records to be accessed selectively without losing alignment.
Example environments from the three profiles are shown in Fig.~\ref{fig:channels4world_at_a_glance}.

This corpus and its core release design were introduced in a preliminary paper~\cite{jang2026channels4world}.
The present article builds on that foundation by expanding the methodology and moving beyond corpus construction to systematic characterization and controlled evaluation.
We organize these extensions around three scientific questions:
(1) What variation is realized across the corpus in environment morphology, material composition, and BS/UE placement?
(2) What multipath information does the packaged corpus retain, and how sensitive is that representation to propagation and path-retention choices?
(3) Under matched training exposure, how does held-out-scene generalization change as environmental breadth increases?

These questions motivate four contributions.
\begin{enumerate}
    \setlength{\itemsep}{0pt}
    \item \textbf{An environment-scale wireless corpus and reproducible procedural scene-generation framework.}
    \mbox{{\sffamily\scshape Channels4World}} contains 100,000 procedurally generated outdoor scenes, each combining an RF environment, dense BS/UE placements, and matched ray-traced multipath records at carrier frequencies of 3.5 and 7~GHz.
    A scene-centered organization preserves the correspondence among these components while allowing environmental breadth and within-environment observation density to be varied separately.
    A common pipeline uses shared coordinate and material conventions, environment-informed placement rules, fixed propagation settings, and automated geometry, placement, and record-integrity checks.
    Indexed components and deterministic nested subsets support selective access and experiments at different scales.
    
    \item \textbf{An empirical characterization and validation of the generated corpus.}
    We establish that all 100,000 scene records are distinct under exact-identity checks, then characterize the realized variation in morphology, RF material composition, and transceiver placement across the three generation profiles.
    To support scene-disjoint learning experiments, we also define a deterministic, profile-stratified 90/10 scene-level train--test split that keeps every scene's geometry, placements, channels at both carrier frequencies, and interaction records together.

    \item \textbf{A characterization and sensitivity analysis of the retained multipath data.}
    We validate the stored multipath records and characterize their behavior across the full corpus, complemented by detailed distributions from a 3,000-scene analysis sample.
    A 48-scene study holds the environments and BS/UE placements fixed while comparing propagation with and without diffraction and testing different path-retention settings.
    The results show that diffraction materially affects retained-link availability and multipath structure, and quantify the tradeoff between retained multipath detail and storage.

    \item \textbf{A controlled evaluation of environmental breadth and generalization.}
    Using nested training sets ranging from 100 to 90,000 scenes under matched training exposure, we compare generalization to held-out scenes as environmental breadth increases.
    Models trained on 10,000 or 90,000 scenes perform better on held-out scenes than the model trained on 1,000 scenes under the tested training budget.
    A matched control holds the number of available episodes fixed but distributes them differently across environments and BS locations.
    The resulting differences are small, indicating that scene count alone does not explain the larger separations in the main training ladder.
    Inference-time controls further show that prediction improves with more context from the same environment and degrades sharply when that context is replaced by context from another environment.
\end{enumerate}

Together, these contributions establish {\sffamily\scshape Channels4World} as both an environment-scale wireless dataset and a framework for studying propagation and learning across environments.
The shared scene structure connects reproducible generation and corpus validation with path-level sensitivity analysis and scene-disjoint generalization, while keeping the two dimensions of wireless-data scale explicit and separately controllable.

The remainder of this paper is organized as follows.
Section~\ref{sec:related_work} reviews existing reusable wireless datasets and related work on spatial channel knowledge, explicit wireless scene construction, and environment-aware wireless learning.
Section~\ref{sec:dataset_design} defines the scene-centered data model, and Section~\ref{sec:generation} presents the procedural generation methodology.
Section~\ref{sec:corpus} characterizes the {\sffamily\scshape Channels4World} corpus, validates record integrity and the canonical public split, and defines the scope of the current corpus.
Section~\ref{sec:evaluation} analyzes retained multipath behavior and sensitivity to propagation and path-retention choices, then evaluates how environmental breadth affects generalization to held-out scenes.
Section~\ref{sec:discussion} discusses limitations, applications, modular extensions, and future research opportunities.
Section~\ref{sec:conclusion} concludes the paper.
The Appendix provides additional details on serialized-record validation, morphology-stratified channel analysis, and the learning representation, training, and controlled comparisons.

\section{Related Work}
\label{sec:related_work}

Wireless data resources differ not only in scale but also in how they represent and preserve the environments from which channel observations are obtained.
Some emphasize configurable channel generation or reusable site-specific observations, while others organize channels as spatial maps, retain explicit scene geometry, or incorporate environmental context into learned representations.
We review these directions with particular attention to the balance between environmental breadth and detail within each environment.
Table~\ref{tab:related_dataset_comparison} compares representative synthetic resources by their environment inventory, sampling within each environment, RF output, retained environmental information, and availability.

\begin{table*}[t]
\caption{Comparison of environment-indexed synthetic wireless resources.}
\label{tab:related_dataset_comparison}
\centering
\footnotesize
\setlength{\tabcolsep}{3pt}
\renewcommand{\arraystretch}{1.12}
\begin{tabularx}{\textwidth}{@{}L{2.15cm}L{1.70cm}L{2.55cm}L{2.50cm}L{1.80cm}YL{1.55cm}@{}}
\toprule
\textbf{Resource} & \textbf{Environment source} & \textbf{Environment inventory / unit} & \textbf{Within-environment sampling} & \textbf{Native RF output} & \textbf{Aligned environment data} & \textbf{Resource availability} \\
\midrule
DeepMIMO v4~\cite{alkhateeb2019deepmimo,deepmimo2026v4} &
Mixed site-specific ray-tracing scenarios &
200+ scenarios in the v4 database; site extent, carrier, and deployment vary &
Scenario-specific Tx/Rx sets; counts and spatial sampling vary by scenario &
Per-path ray-tracing data; configurable MIMO channels computed on demand &
Tx/Rx positions and path data; geometry, materials, and system metadata where provided &
Public v4 toolchain and scenario database \\
\addlinespace[1pt]

WAIR-D~\cite{huangfu2022waird} &
Geographic import (OSM, 40+ cities) &
10,000 sparse environments; 100 dense environments &
Sparse: 5 BS $\times$ 30 UE. Dense: 1 BS $\times$ 10,000 UE &
Ray-traced channels &
Environment image, BS/UE positions, and region size &
Public dataset \\
\addlinespace[1pt]

RadioMapSeer~\cite{yapar2022radiomapseer} &
Geographic import (OSM, 6 cities) &
701 maps, each $256{\times}256$ m &
80 Tx/map on a 1-m receiver grid &
Path-loss radio maps &
OSM city maps, roads, Tx locations, and simulator variants &
Public dataset \\
\addlinespace[1pt]

U6G XL-MIMO~\cite{li2026u6gradiomap} &
Geographic import (OSM, Nanjing) &
800 scenes, each $1.28{\times}1.28$ km &
98 Tx configurations/scene on a $128{\times}128$ receiver grid at 10-m spacing &
78,400 configuration-specific beamformed received-power maps &
Height and beam maps, separately available meshes, and fixed material roles: marble buildings, metallic roofs, and concrete ground &
Public data and code \\
\addlinespace[1pt]

Wi3R / WiPTR~\cite{hehn2025wigatr} &
Procedural generation (indoor) &
5,000 / 12,000 scenes &
5 Tx $\times$ 200 Rx / 1--15 Tx and up to 200 Rx &
Per-path channels (25 strongest paths) &
3D meshes, fixed role-based materials (2 / 6 classes), and transceiver placements &
Research-use release \\
\addlinespace[1pt]

OpenPathNet~\cite{liu2025openpathnet} &
Geographic import (OSM) with procedural fallback &
Paper: $>$12,000 $128{\times}128$ m scenes. Releases: $\sim$1,700 curated / $\sim$10,000 larger &
1 Tx on a $128{\times}128$ receiver grid at 1-m spacing &
Per-path gain, delay, angles, and type &
Scene XML and meshes with assigned materials, placements, and path records; scene-level material variation not characterized &
Public generator and two releases \\
\midrule

Channels4World (this work) &
Procedural generation (outdoor) &
100,000 scenes, each with one procedurally generated environment in a uniform $500{\times}500$ m domain &
32 BS and 16,384 UE positions per scene, evaluated at both 3.5 and 7~GHz &
Variable-length multipath records with up to 16 retained paths per link &
Geometry with configured per-building material variation, placements, frequency-specific channel records at 3.5 and 7~GHz, and ordered interaction records &
Public via project website \\
\bottomrule
\end{tabularx}
\vspace{1pt}

\parbox{\textwidth}{\scriptsize\textit{Note:} Inventory units follow each resource's published organization; database scenarios, source environments, maps, and configuration-specific maps are not treated as interchangeable counts. Dynamic catalog and release information reflects versions accessed in September 2026.}
\end{table*}

\subsection{From Channel Models to Reusable Wireless Datasets}

Configurable channel models remain a central source of controlled wireless data.
The 3GPP TR~38.901 model provides standardized scenario-dependent channel generation across frequencies from 0.5 to 100~GHz~\cite{3gpp38901}.
COST 2100 models multipath as evolving clusters governed by visibility regions, while QuaDRiGa extends geometry-based stochastic modeling to three-dimensional multicell channels with time evolution~\cite{liu2012cost2100,jaeckel2014quadriga}.
Both support sampling under specified scenario and mobility assumptions.
Their primary object, however, is a parameterized channel law.
Reusable datasets serve a different role by providing common observations tied to explicit sites, positions, and system assumptions.

Early ray-tracing studies used site-specific 5G channel data for learning-based beam selection~\cite{Klautau18}.
DeepMIMO later introduced a reusable, parameterized framework for constructing machine-learning-ready multiple-input multiple-output (MIMO) channels from site-specific ray-tracing data~\cite{alkhateeb2019deepmimo}.
Its channel matrices retain their dependence on environment geometry, materials, and transmitter/receiver locations, while configurable system parameters allow users to derive task-specific MIMO datasets from a shared ray-tracing scenario.
ViWi pairs synthetic channels with aligned visual observations~\cite{alrabeiah2019viwi}, and Raymobtime adds mobility while maintaining consistency across time, space, and frequency~\cite{klautau2021raymobtime}.
Together, these resources establish three useful principles: preserve site context, state system assumptions, and retain aligned modalities or trajectories when the learning task depends on them.

Later datasets broaden either site coverage or modality coverage.
WAIR-D provides sparse sampling across 10,000 OpenStreetMap (OSM)-derived environments and dense sampling across 100 environments~\cite{huangfu2022waird}.
DataAI--6G supports channel generation under configurable system parameters, whereas M3SC and Multimodal-Wireless align simulated channels with visual, depth, LiDAR, radar, and other sensing data~\cite{shen2023dataai6g,cheng2023m3sc,mao2026multimodalwireless}.
DeepSense 6G instead provides synchronized measurements from real communication and sensing hardware across more than 40 deployment scenarios~\cite{alkhateeb2023deepsense}.
Synthetic resources can support controlled variation and exact alignment among generated modalities, whereas measured resources capture hardware and propagation effects that simulation may omit.

\subsection{Radio Maps and Spatial Channel Knowledge}

Building on site-indexed channel observations, dense spatial sampling organizes propagation quantities over an environment.
RadioUNet introduced fast geometry-conditioned path-loss-map prediction and the RadioMapSeer training corpus~\cite{levie2021radiounet}.
A later dataset release built on the same 701 urban layouts adds time-of-arrival maps for localization, with 80 transmitters evaluated on a common receiver grid~\cite{yapar2022radiomapseer}.
Geo2SigMap automates a workflow from geographic data through Sionna simulation to signal-map prediction, then evaluates the model with measurements~\cite{li2023geo2sigmap}.
Channel knowledge maps provide a broader abstraction for organizing location-dependent propagation information in environment-aware communication and sensing~\cite{zeng2024ckm}.
Together, these approaches make spatial continuity and environmental context explicit parts of the data representation.

Recent datasets extend radio maps beyond two-dimensional, single-band path loss.
SpectrumNet spans multiple frequencies, receiver heights, terrain types, and climate conditions~\cite{zhang2025spectrumnet}.
CKMImageNet pairs location-specific channel knowledge with environmental maps and visual representations~\cite{wu2025ckmimagenet}.
RadioDiff-3D introduces UrbanRadio3D, with volumetric path-loss, direction-of-arrival, and time-of-arrival fields plus diffusion-based reconstruction benchmarks~\cite{wang2025radiodiff3d}.
UrbanMIMOMap supplies dense complex MIMO channel state information (CSI) from which precoding-aware maps can be derived~\cite{jia2025urbanmimomap}.
U6G applies 98 transmitter configurations across 800 urban scenes and five bands to produce 78,400 configuration-specific received-power maps~\cite{li2026u6gradiomap}.
These examples show how one environment can support many receiver, carrier, and array configurations.

\subsection{Explicit Geometry and Wireless Scene Construction}

Beyond spatial organization of channel quantities, geometry-aware simulation makes the physical environment an explicit, reusable input rather than an unobserved source of channel samples.
Sionna RT provides an open, GPU-accelerated, differentiable ray tracer with path and radio-map solvers over three-dimensional scenes~\cite{hoydis2023sionnart,aoudia2025sionnart}.
WiSegRT pairs segmented indoor geometry with site-specific channel impulse responses~\cite{zhang2024wisegrt}, while Wi-GATr learns propagation from scene primitives and evaluates transfer to unseen layouts~\cite{hehn2025wigatr}.
The associated Wi3R and WiPTR corpora retain thousands of indoor layouts, transceiver placements, channel observations, and meshes.
These resources show how a scene can remain a reusable scientific object rather than serving only as an intermediate input to a derived channel representation.

Outdoors, scaling explicit scenes requires a systematic source of complete environments.
Geographic import is one route.
RadioMapSeer and Geo2SigMap use geographic data to construct site-indexed radio maps, while Sensiverse preserves acquired three-dimensional scene models, channel data, and transmitter/receiver associations for integrated sensing and communication (ISAC) studies~\cite{yapar2022radiomapseer,li2023geo2sigmap,luo2023sensiverse}.
OpenPathNet provides a path-level precedent.
Its open OSM/Blender/Sionna pipeline constructs Sionna-compatible scenes, validates receiver locations against geometry, and exports per-path gain, delay, angles, and type with the scene assets~\cite{liu2025openpathnet}.
Geographic import anchors layouts in real places and can preserve their site-specific structure.

Procedural generation offers a second route, creating scene variation under specified rules.
Li \emph{et al.} use an ITU-R-parameterized virtual city as a ray-tracing basis for directional-channel modeling, while Panda5gSim integrates synthetic urban layouts with ground and aerial mobility~\cite{li2019directional,raddwan2024panda5gsim}.
These examples highlight choices between imported and generated geometry and between random and deployment-aware placement.
The remaining challenge is to assemble many complete environments with dense, aligned observations, documented provenance, and clear evaluation splits at a scale suitable for pretraining reusable models.

\subsection{Wireless Foundation Models and Environment-Scale Data}

With channel observations increasingly tied to spatial and scene context, wireless foundation models raise a complementary question: which variations must a pretraining corpus cover to support reuse?
The design space is still heterogeneous and evolving, and recent survey work has begun to organize these efforts under the broader channel-foundation-model perspective~\cite{jiang2025cfm}.
LWM learns task-agnostic channel embeddings through masked channel modeling, while WiFo uses masked space--time--frequency reconstruction for zero-shot channel prediction across heterogeneous CSI configurations~\cite{alikhani2025lwm,liu2025wifo}.
WiMAE uses masked reconstruction, while ContraWiMAE extends it with masked contrastive learning for transfer across discriminative and reconstruction tasks~\cite{guler2025multitask}.
CSI-CLIP treats channel impulse responses and CSI as paired views for contrastive alignment~\cite{jiang2025csiclip}.
WirelessGPT was introduced as a generative pre-trained multi-task framework for wireless communication and later formulated as a generative foundation model for multi-task integrated sensing and communication~\cite{yang2025wirelessgpt,yang2026wirelessgpt}.
WavesFM shares a backbone across spectrogram, CSI, and in-phase/quadrature (I/Q) inputs, while IQFM uses contrastive pretraining directly on measured multi-antenna I/Q streams~\cite{aboulfotouh2025wavesfm,mashaal2025iqfm}.
Across these models, differences in input representation, pretraining objective, adaptation protocol, and intended transfer scope shape the variation that their pretraining data must cover.

A second line of work makes environmental context itself part of the learned representation.
MUSE-FM combines a unified multitask architecture with scene-graph context~\cite{zheng2025musefm}.
Wireless World Model fuses CSI, three-dimensional point clouds, and trajectories to predict spatiotemporal channel evolution~\cite{chen2026wirelessworldmodel}.
Telecom World Models provides a broader framing that connects digital twins, foundation models, and predictive planning for telecommunications~\cite{zou2026telecomworldmodels}.
These models incorporate environmental knowledge through explicit modalities.
Channel2World takes a different route by treating the RF environment as the central object of representation learning.
It infers a wireless world embedding from position-indexed multipath observations within a BS-centered environment and reuses that embedding as a prior across downstream tasks~\cite{moon2026channel2world}.
These directions motivate preserving scene identity and using scene-level evaluation boundaries when testing generalization to unseen environments.

Together, these approaches make the environment itself a meaningful unit for representation and evaluation.
Environment awareness, however, does not by itself establish generalization beyond settings seen during training.
Existing resources increasingly span multiple settings and preserve scene context, but the effect of training-environment breadth on held-out-scene performance remains largely uncharacterized.
In existing datasets and experiments, environmental breadth, within-environment observation density, total data exposure, and representation choices are often coupled, making their individual effects difficult to isolate.
{\sffamily\scshape Channels4World} makes environmental breadth and within-environment observation density separately controllable across 100,000 outdoor scenes with dense, aligned multipath observations.
This scale supports both detailed within-scene analysis and controlled evaluation of how broader training coverage affects generalization.

\section{Scene-Centered Dataset Design}
\label{sec:dataset_design}

An \emph{environment} is the physical propagation setting defined by its geometry and RF material assignments.
A \emph{scene} is the corpus unit that binds one environment to one BS/UE placement set and the resulting channel and interaction records at 3.5 and 7~GHz.
This shared scene identity anchors the corresponding frequency-specific evaluations.

\subsection{Environment and Placement Structure}

We use \(s\), \(b\), \(u\), and \(f\) to index the scene, BS, UE, and carrier, respectively.
Let \(\mathcal{F}=\{3.5~\mathrm{GHz},\,7~\mathrm{GHz}\}\), with \(f\in\mathcal{F}\).
The environment component is
\begin{equation}
\mathcal{E}_s=(\mathcal{G}_s,\mathcal{M}_s),
\end{equation}
where \(\mathcal{G}_s\) is the three-dimensional geometry and \(\mathcal{M}_s\) contains its material assignments.
The placement component is
\begin{align}
\mathcal{X}_s
&=
\left(\mathbf{X}^{\mathrm{B}}_s,\mathbf{X}^{\mathrm{U}}_s\right),\\
\mathbf{X}^{\mathrm{B}}_s
&=
\left[
\mathbf{x}^{\mathrm{B}}_{s,1},\ldots,
\mathbf{x}^{\mathrm{B}}_{s,N_B}
\right]^{\mathsf T}
\in\mathbb{R}^{N_B\times3},\\
\mathbf{X}^{\mathrm{U}}_s
&=
\left[
\mathbf{x}^{\mathrm{U}}_{s,1},\ldots,
\mathbf{x}^{\mathrm{U}}_{s,N_U}
\right]^{\mathsf T}
\in\mathbb{R}^{N_U\times3}.
\end{align}
Here, \(\mathbf{x}^{\mathrm B}_{s,b}\) and \(\mathbf{x}^{\mathrm U}_{s,u}\) are BS and UE coordinates in the common metric Cartesian frame of environment \(\mathcal{E}_s\).
The corpus uses \(N_B=32\) and \(N_U=16{,}384\).

For each scene, the environment geometry, material assignments, and transceiver positions are held fixed while every BS--UE pair is evaluated separately at 3.5 and 7~GHz.
This isolates frequency-dependent propagation from changes in the physical environment or deployment.
We denote one frequency-specific evaluation by \(q=(s,b,u,f)\).
Table~\ref{tab:link_hierarchy} provides the corresponding unit and key mapping.
The two evaluations for a given physical BS--UE pair share the same environment and placement variables, but their retained paths need not correspond one-to-one because propagation strength and retention outcomes can differ across frequencies.

\subsection{Link Evaluations and Retained Paths}

Every BS--UE pair is evaluated once at each carrier before applying the retention rule.
Each frequency-specific evaluation produces ray-tracing output containing zero or more candidate propagation paths.
An evaluation with at least one path surviving the retention rule is a retained link, and each surviving candidate becomes a retained path.

\begin{table}[t]
\caption{Link and path hierarchy.}
\label{tab:link_hierarchy}
\centering
\footnotesize
\setlength{\tabcolsep}{3pt}
\renewcommand{\arraystretch}{1.10}
\begin{tabularx}{\columnwidth}{@{}L{0.24\columnwidth}L{0.22\columnwidth}Y@{}}
\toprule
\textbf{Unit} & \textbf{Key} & \textbf{Meaning}\\
\midrule
BS--UE pair & \((s,b,u)\) & One BS--UE position pair in scene \(s\), shared by both frequencies\\
Frequency-specific evaluation & \(q=(s,b,u,f)\) & That pair evaluated at one frequency\\
Retained link & \(q\) with \(K_q>0\) & An evaluation with at least one retained path\\
Retained path & \((q,k)\) & The \(k\)th power-ordered path of a retained link\\
\bottomrule
\end{tabularx}
\end{table}

For each evaluation \(q\), the retention rule produces the possibly empty, power-ordered retained-path sequence
\begin{equation}
\mathcal{P}_{q}
=
\left(
p_{q,1},\ldots,
p_{q,K_q}
\right),
\qquad
|a_{q,1}|^2\geq\cdots\geq|a_{q,K_q}|^2.
\end{equation}
Here, \(K_q\) is the number of paths retained for evaluation \(q\), and \(a_{q,k}\) is the complex coefficient of the \(k\)th retained path.
When \(K_q=0\), \(\mathcal{P}_q\) is empty.
Section~\ref{sec:generation} gives the exact ranking, cap, and power threshold.

Each scene contains 524,288 BS--UE pairs, and evaluating every pair at both carriers produces two evaluations per pair before applying the retention rule.
Section~\ref{sec:corpus} reports the corpus-wide evaluation, retained-link, and retained-path counts together with their denominators.

\subsection{Path-Level Channel and Interaction Representation}

{\sffamily\scshape Channels4World} preserves individual propagation paths rather than committing each link to one bandwidth, subcarrier grid, array-specific channel tensor, or learning representation.
For retained path \((q,k)\), with \(k\in\{1,\ldots,K_q\}\), the retained-path descriptor is
\begin{equation}
p_{q,k}
=
\left(
a_{q,k},
\tau_{q,k},
\boldsymbol{\Omega}^{\mathrm{D}}_{q,k},
\boldsymbol{\Omega}^{\mathrm{A}}_{q,k},
J_{q,k},
\boldsymbol{\eta}^{\mathrm{sum}}_{q,k}
\right).
\end{equation}
Here, \(a_{q,k}\in\mathbb{C}\) is the dimensionless complex passband path coefficient under the simulation antenna convention, so \(|a_{q,k}|^2\) is the path power gain, and \(\tau_{q,k}\) is the absolute propagation delay.
\(\boldsymbol{\Omega}^{\mathrm{D}}_{q,k}\) and \(\boldsymbol{\Omega}^{\mathrm{A}}_{q,k}\) are the departure and arrival zenith--azimuth pairs.
These angles use the shared right-handed scene frame, with zenith measured from \(+z\) and azimuth measured in the \(x\)--\(y\) plane from \(+x\) toward \(+y\).
Departure is defined at the BS and arrival at the UE.
The remaining quantities encode the interaction count \(J_{q,k}\) and interaction-summary flags \(\boldsymbol{\eta}^{\mathrm{sum}}_{q,k}\).
Because \(\mathcal{P}_{q}\) is ordered by descending power, \(k\) also gives the power rank within the retained sequence.

To connect the path-level corpus format with a conventional scalar channel formulation, consider a retained link \(q\) with \(K_q>0\).
Let \(f_q\) denote the carrier frequency associated with evaluation \(q\), and define \(a^{\mathrm{b}}_{q,k}=a_{q,k}e^{-j2\pi f_q\tau_{q,k}}\) as the corresponding baseband-equivalent path coefficient.
The retained coefficient--delay pairs then define the baseband channel impulse response (CIR) and channel frequency response (CFR)
\begin{align}
h_q(t)
&=
\sum_{k=1}^{K_q}
a^{\mathrm{b}}_{q,k}\delta\!\left(t-\tau_{q,k}\right),\\
H_q(\nu)
&=
\sum_{k=1}^{K_q}
a^{\mathrm{b}}_{q,k}e^{-j2\pi\nu\tau_{q,k}},
\end{align}
where \(\nu\) is the baseband frequency offset from the carrier associated with evaluation \(q\).
These expressions provide a scalar representation of the retained channel in the time and frequency domains.
Once a bandwidth and sampling convention are specified, these expressions can be evaluated on the desired time or frequency grid.
The departure and arrival angles describe the spatial directions of the same retained paths, which the scalar formulation does not represent.

For later channel summaries, we define the linear retained coefficient-power sum as
\begin{equation}
S_q=\sum_{k=1}^{K_q}|a_{q,k}|^2.
\end{equation}
When reported in decibels, the corresponding quantity is \(10\log_{10}S_q\), which we call the retained coefficient-power sum in dB.

The retained path coefficients and angles support array-dependent MIMO representations under compatible element-pattern and polarization conventions.
These representations can be used to evaluate beamforming, combining, beam selection, and precoding.
The dataset schema and validation analysis specify the coefficient normalization, units, and other conventions needed to interpret these quantities.

For each retained path \((q,k)\), its ordered interactions form the BS-to-UE sequence
\begin{equation}
d_{q,k}
=
\left[
\left(\kappa_{q,k,j},\mathbf{v}_{q,k,j}\right)
\right]_{j=1}^{J_{q,k}}.
\end{equation}
A line-of-sight (LoS) path has \(J_{q,k}=0\) and an empty \(d_{q,k}\).
Otherwise, \(\boldsymbol{\eta}^{\mathrm{sum}}_{q,k}\) indicates which interaction mechanisms occur, while \(d_{q,k}\) preserves their BS-to-UE order and scene-frame vertices.
Each pair in this sequence is an interaction record.
Here, \(\kappa_{q,k,j}\) identifies the propagation mechanism and \(\mathbf{v}_{q,k,j}\in\mathbb{R}^3\) gives its position in the shared scene frame.
The ordered interaction sequence for evaluation \(q\) is
\begin{equation}
\mathcal{D}_q
=
\left(
d_{q,1},\ldots,
d_{q,K_q}
\right).
\end{equation}
The sequences \(\mathcal{D}_q\) and \(\mathcal{P}_q\) follow the same descending-power path order, so \(d_{q,k}\) and \(p_{q,k}\) describe the same retained path.
This separation makes physical path trajectories available for path-geometry studies while keeping the retained-path descriptor limited to coefficients, delays, angles, interaction counts, and summary flags.

For scene \(s\), the collection of frequency-specific evaluations is
\begin{align}
\mathcal{Q}_s
=
\{(s,b,u,f):{}& 1\leq b\leq N_B,\ 1\leq u\leq N_U,\nonumber\\
&f\in\mathcal{F}\}.
\end{align}
At the logical dataset level, one scene-indexed record is
\begin{equation}
\mathcal{R}_s
=
\left(
\mathcal{E}_s,
\mathcal{X}_s,
(\mathcal{P}_q)_{q\in\mathcal{Q}_s},
(\mathcal{D}_q)_{q\in\mathcal{Q}_s}
\right).
\end{equation}
Both collections are indexed by the same set \(\mathcal{Q}_s\).
For each evaluation \(q\), \(\mathcal{P}_q\) and \(\mathcal{D}_q\) contain \(K_q\) records in the same descending-power path order.
When \(K_q=0\), both are empty sequences in this logical representation.
Detailed metadata, field schemas, and physical file organization are specified in the dataset documentation.
Section~\ref{sec:generation} describes how these aligned components are generated and validated.

\section{Procedural Environment and Channel Generation}
\label{sec:generation}

\begin{figure*}[t]
\centering
\includegraphics[width=\textwidth]{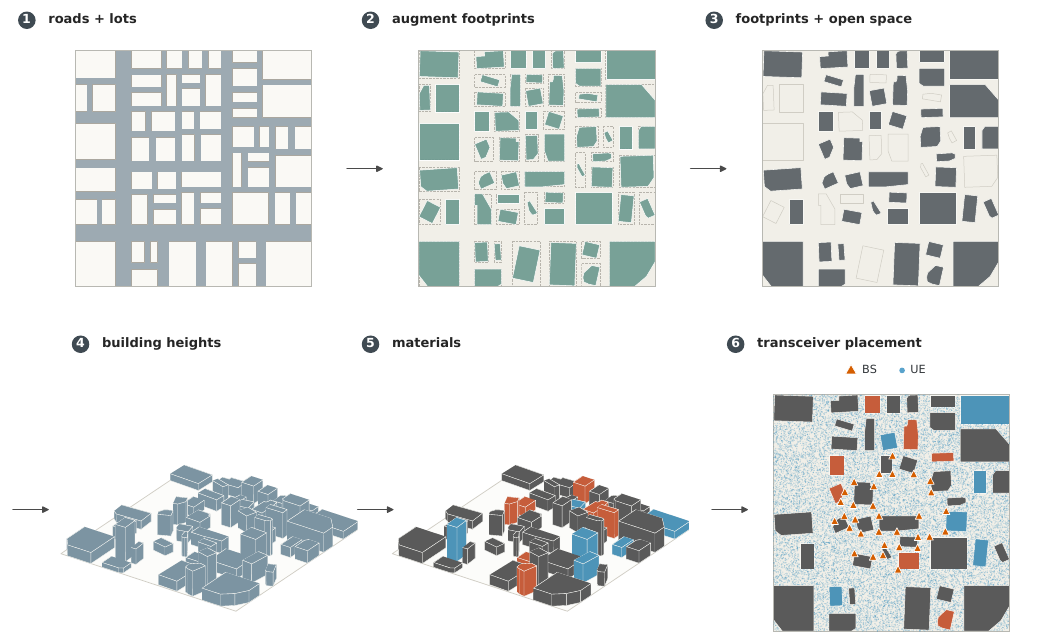}
\caption{Procedural environment construction and environment-informed transceiver placement.}
\label{fig:procedural_construction_and_placement}
\end{figure*}

For each scene index \(s\), procedural environment construction and placement produce \(\mathcal{E}_s\) and \(\mathcal{X}_s\), which are validated before simulation.
For every \(q \in \mathcal{Q}_s\), propagation simulation yields a candidate-path sequence \(\mathcal{C}_q\), and the retention rule converts its surviving candidates into the aligned \(\mathcal{P}_q\) and \(\mathcal{D}_q\) sequences.
Once all required components are complete and the integrity checks pass, they are assembled into the scene-indexed record \(\mathcal{R}_s\).
Fig.~\ref{fig:procedural_construction_and_placement} provides an overview of the environment-construction and placement stages.

Rather than model fine visual detail, the method prioritizes propagation-relevant structure at corpus scale: road-scale open space, building massing and footprint shape, height, material class, and environment-informed transceiver placement.
The same staged procedure is used for the Urban, Metropolitan, and Suburban generation profiles, with profile-specific settings controlling road and lot structure, building morphology, and material composition.

\subsection{Environment Geometry and Material Assignment}

\subsubsection{Road Layout and Source Lots}
Generation begins with the centered \(500\times500\)~m environment domain.
Depending on the generation profile, major road corridors may first divide the land into primary blocks, which are then recursively subdivided by progressively smaller corridors.
The remaining polygons form source lots.
The three profiles vary major-road structure, road spacing and width, recursion, and lot dimensions.
Roads are represented as negative space within the ground plane rather than as separate raised meshes.

\subsubsection{Building Footprints and Open Space}
The source lots produced by the road stage define the parcels within which buildings may be placed, but using each lot boundary directly as a building footprint would make building shape follow the road subdivision too closely.
Instead, each lot serves as an envelope for a candidate footprint.
The generator first creates a smaller footprint within the lot with randomized size and orientation and may further modify its corners with diagonal cuts or rectangular notches to introduce nonrectangular shapes.
Each accepted transformation must remain inside the source lot, preserving the larger-scale road and block structure while adding local geometric variation.
Some candidate lots are intentionally left unbuilt to create open space.
The remaining footprints are checked for valid, hole-free geometry and sufficient area, and overly elongated shapes are rejected before the accepted footprints become buildings.

\subsubsection{Building Heights, Meshes, and Materials}
After the building footprints are selected, each is assigned a height.
Heights are sampled from profile-dependent log-normal distributions whose parameters also depend on footprint area.
The height model uses a scene-level reference footprint size: buildings with smaller footprints can be taller, while those with very large footprints tend to be shorter.
Heights still vary randomly among buildings with similar footprint areas, and a footprint-dependent cap prevents excessively tall, narrow structures.
Each accepted footprint is then extruded vertically into a closed, flat-roof mesh.

Material assignment is performed after the geometry is fixed and does not change the building shapes.
Each building is assigned one RF material according to profile-specific probabilities.
Concrete is the most common material in all three profiles; Urban and Metropolitan also include brick and glass, whereas Suburban includes brick and wood.
Because materials are assigned building by building, individual scenes contain material variation, while the profile-specific probabilities shift the overall composition across profiles.
The ground is concrete throughout.
The generation settings record these probabilities, and Section~\ref{sec:corpus} reports the material shares realized in the full corpus.

Together, the resulting geometry and material assignments form the environment component
\(\mathcal{E}_s=(\mathcal{G}_s,\mathcal{M}_s)\).

\subsection{Environment-Informed BS and UE Placement}

Once the environment geometry and materials are fixed, BS and UE positions are generated from the completed scene.
BS placement is tied to building geometry, whereas UEs are sampled over ground space outside buildings.
Neither placement stage modifies the accepted environment.

\subsubsection{BS Placement}
BS candidates are constructed from building roof vertices, with preference given to vertices nearer the center of the environment.
This center-oriented selection reduces the number of BSs near the finite scene boundary, where structures outside the modeled domain are absent.
Within each building, spacing constraints and a per-building cap limit how many candidates can be selected, and a final global spacing rule distributes the chosen BSs across the scene.

Each selected roof vertex is then displaced horizontally outward from the building boundary so that the final BS position lies outside the building volume.
Its vertical coordinate is determined from the roof height and a 15--50~m deployment-height band.
Depending on the building height, the resulting BS may lie at rooftop level, beside the facade of a taller building, or elevated above a low roof.
Every final position is checked against all building volumes and rejected if it lies inside one.

\subsubsection{UE Placement and Clearance}
UE positions are sampled from ground space outside the accepted building footprints.
To generate candidates efficiently, the completed footprints are first rasterized into a binary mask that distinguishes building-occupied cells from the remaining ground area.
The mask does not assign semantic labels such as road, sidewalk, or plaza; an allowed cell simply indicates ground that is not occupied by a building.
Because the raster mask is only a proposal mechanism, final validity is checked against the continuous building geometry.

\begin{samepage}
For a horizontal UE position \(\mathbf{x}\in\mathbb{R}^{2}\) and closed building footprints \(\{B_i\}\), validity requires
\begin{equation}
\min_i \operatorname{dist}(\mathbf{x},B_i)\geq d_{\min},
\qquad
d_{\min}=0.5\ \text{m}.
\end{equation}
\end{samepage}
Here, \(\operatorname{dist}\) is the Euclidean point-to-set distance.
An allowed raster cell is sampled with replacement, a horizontal coordinate is drawn uniformly within that cell, and UE height is sampled uniformly from 1 to 2~m.
The continuous-geometry check rejects positions that fall inside, on, or within \(d_{\min}\) of any building footprint.
Sampling continues until all \(N_U\) valid UE positions are obtained or the fixed proposal limit is reached.

The resulting BS and UE coordinate arrays populate \(\mathbf{X}^{\mathrm{B}}_s\) and \(\mathbf{X}^{\mathrm{U}}_s\), respectively, and together form the placement component \(\mathcal{X}_s\).

\subsection{Deterministic Sampling and Pre-Simulation Validation}

\subsubsection{Domain-Separated Random Streams}
Although environment construction and placement contain random choices, each scene is generated reproducibly from a stored base seed.
Rather than allowing every stage to consume one shared random-number sequence, the generator derives separate streams for road layout, footprint shaping, open-space and filtering decisions, heights, materials, BS placement, and UE placement.

For scene \(s\), semantic domain \(d\), and substream index \(i\),
\begin{equation}
\sigma_{s,d,i}
=
\operatorname{Hash}(\sigma_s\mathbin{\|}d\mathbin{\|}i),
\qquad
R_{s,d,i}
=
\operatorname{PRNG}(\sigma_{s,d,i}),
\end{equation}
where \(\sigma_s\) is the stored base seed for scene \(s\), \(d\) identifies the generation stage, and \(i\) identifies a reproducible substream within that stage.
The operator \(\mathbin{\|}\) denotes deterministic composition of these identifiers, while \(\operatorname{Hash}\) and \(\operatorname{PRNG}\) denote the specified seed-derivation and pseudorandom-generation operations.

This separation limits how changes in one stage alter the random draws used by other stages and supports deterministic regeneration of the environment and placements from the stored seed, settings, and generation state.
It also makes individual stages easier to inspect, modify, and ablate.
Domain separation concerns random-number consumption rather than statistical independence: changes to the generated geometry can still alter the inputs available to downstream stages.

\subsubsection{Pre-Simulation Validation}
Before ray tracing, each candidate environment and placement pair
\((\mathcal{E}_s,\mathcal{X}_s)\) must pass a common set of automated checks.
These checks verify that the building footprints and meshes are valid, that the environment and placement records are mutually consistent, and that the BS and UE position arrays are complete and finite.
They also enforce the placement constraints introduced above: BS positions must remain outside building interiors, and UE positions must satisfy the specified building-clearance requirement.

Candidates that fail any of these checks are rejected before simulation and replaced automatically.
The Suburban profile additionally applies a deterministic placement-quality check that limits excessive BS concentration near the finite environment boundary.
Section~\ref{subsec:scene_completion} describes the subsequent replacement and final corpus-admission procedure.

\subsection{Ray-Traced Propagation Simulation}

Each validated scene is simulated separately at 3.5 and 7~GHz using Sionna RT, while the environment geometry and transceiver placements remain fixed across the two carriers~\cite{hoydis2023sionnart,aoudia2025sionnart}.
For every frequency-specific evaluation \(q\in\mathcal{Q}_s\), the solver produces a set of candidate propagation paths under the common settings summarized in Table~\ref{tab:propagation_config}.
Building material labels use the ITU-R P.2040-3 models implemented by the specified Sionna RT release, so their relative permittivity and conductivity are evaluated at the active carrier~\cite{itu2023p2040}.

\begin{table}[t]
\caption{Fixed simulation and path-retention settings.}
\label{tab:propagation_config}
\centering
\footnotesize
\setlength{\tabcolsep}{4pt}
\renewcommand{\arraystretch}{1.08}
\begin{tabularx}{\columnwidth}{@{}L{0.35\columnwidth}Y@{}}
\toprule
\textbf{Property} & \textbf{Fixed setting}\\
\midrule
Direction and carriers & Downlink at 3.5 and 7~GHz\\
Antenna convention & Single-element isotropic V-polarized transmit and receive antennas\\
Modeled mechanisms & Line of sight (LoS), specular reflection, and diffraction with the lit region enabled\\
Mechanisms not modeled & Diffuse scattering, refraction or transmission, and edge diffraction\\
Maximum interaction depth & 4\\
Material model & ITU-R P.2040-3 classes with frequency-dependent properties\\
Surface controls & 0.1-m thickness, zero scattering coefficient, and zero cross-polarization discrimination (XPD) coefficient\\
Candidate-path search & 262,144 samples/transmitter for Urban and Metropolitan; 524,288 samples/transmitter for Suburban; fixed solver seed 42\\
Path retention & Top \(K\) candidates by \(|a|^2\), with \(K=16\), followed by \(10\log_{10}|a|^2>-150\)~dB\\
Simulation stack & Sionna RT 2.0.1, Mitsuba 3.8.0~\cite{jakob2022mitsuba3}, and Dr.Jit 1.3.1~\cite{jakob2022drjit}\\
\bottomrule
\end{tabularx}
\end{table}

The propagation model includes line of sight (LoS), specular reflection, and diffraction, with paths traced to a maximum interaction depth of four.
The separate edge-diffraction option for unshared mesh boundaries is disabled to avoid contributions from artificial finite boundaries such as the outer edge of the ground plane~\cite{aoudia2025sionnart}.
Diffuse scattering is also disabled because the procedural scenes do not provide surface-specific scattering calibration.
Refraction and transmission are omitted because the buildings are represented as exterior shells without calibrated interiors, floors, openings, or indoor materials.
These choices define the propagation scope of the released channel data rather than implying that the omitted mechanisms are physically unimportant.

Candidate paths are discovered using Sionna's seeded shooting-and-bouncing search.
During production, each solver call contains 16 transmitters and 1,024 receivers, with the sampling budgets and fixed solver seed given in Table~\ref{tab:propagation_config}.

Before path retention is applied, the ray tracer returns for each evaluation \(q\) the candidate-path sequence
\begin{equation}
\mathcal{C}_q
=
(c_{q,1},\ldots,c_{q,M_q}),
\qquad
M_q = |\mathcal{C}_q|.
\end{equation}
Here, \(M_q\) is the number of solver-returned candidate paths available to the retention stage, not the number of rays considered internally during the search, and \(a_{q,m}\) denotes the complex coefficient of candidate \(m\).
The full production campaign used four NVIDIA GeForce RTX 5090 GPUs and spanned roughly two weeks, including interruptions, recovery, and finalization.

\subsection{Path Retention by Power Rank and Threshold}

For each evaluation, candidate paths are first ordered by coefficient power.
Up to \(K=16\) of the strongest candidates are considered for retention, and only those whose power gain exceeds the strict \(-150\)-dB threshold are kept.
Formally, for the candidate sequence \(\mathcal{C}_q\), let \(\pi_q(j)\) denote the candidate index at descending-power rank \(j\).
Power ties preserve the existing deterministic candidate order.
With \(K=16\) and \(\gamma=-150\)~dB, the surviving candidate ranks are
\begin{equation}
\begin{aligned}
\mathcal{I}_q=\bigl\{\, & j\in\{1,\ldots,\min(K,M_q)\}: \\
& 10\log_{10}\!\left|a_{q,\pi_q(j)}\right|^2>\gamma\,\bigr\}.
\end{aligned}
\end{equation}
The inequality is strict.
Each surviving candidate is converted into the retained-path descriptor \(p_{q,k}\) together with its aligned interaction sequence \(d_{q,k}\), preserving descending-power order.
These records form the \(\mathcal{P}_q\) and \(\mathcal{D}_q\) sequences defined in Section~\ref{sec:dataset_design}.
The retained-path count is therefore \(K_q=|\mathcal{I}_q|\), with
\(0\leq K_q\leq\min(K,M_q)\).
The power threshold is a storage criterion rather than a receiver-sensitivity threshold.
Whether a retained link supports communication depends on the transmit power and receiver assumptions chosen for the downstream task.

\begin{figure*}[t]
\centering
\includegraphics[width=\textwidth]{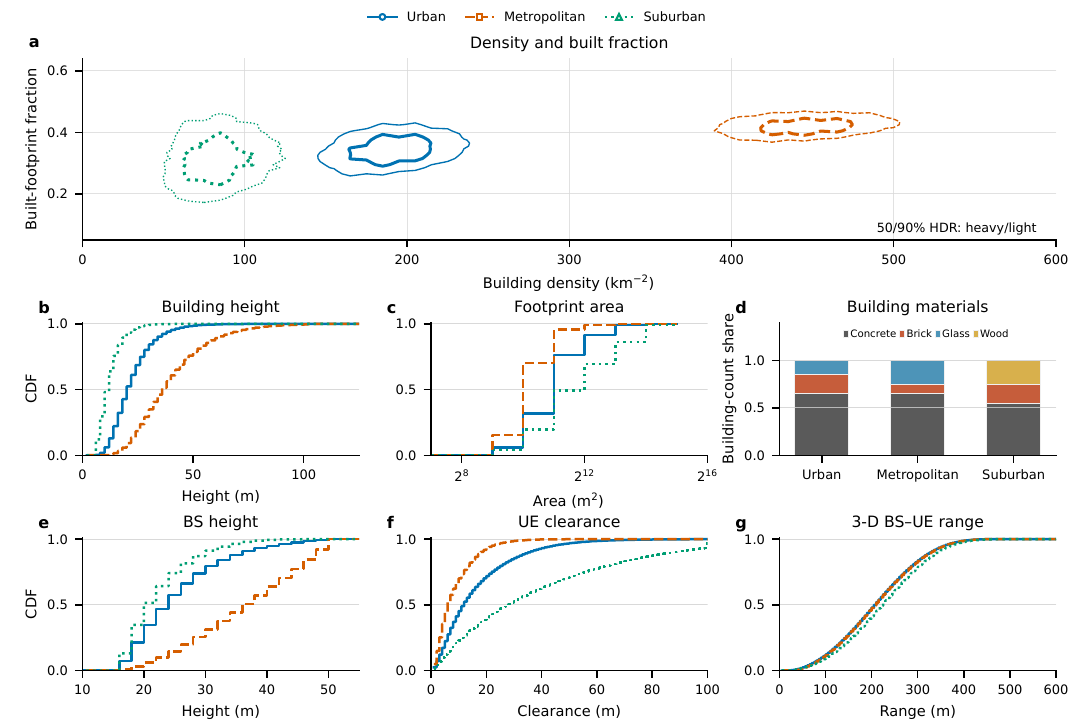}
\caption{Environment and placement distributions across the three generation profiles. In panel (a), the heavy and light contours enclose the 50\% and 90\% highest-density regions (HDRs) of the joint distribution, respectively.}
\label{fig:environment_placement_characterization}
\end{figure*}

\subsection{Scene Completion and Corpus Admission}
\label{subsec:scene_completion}

A candidate scene enters the corpus only after its environment, placements, and channel and interaction records at both frequencies are complete and all file-integrity checks have passed.
This all-or-nothing admission rule prevents partial or single-frequency scene records from entering the release.
If generation, simulation, validation, or export fails and recovery is not possible, the candidate is discarded and a replacement is generated from the next candidate seed for the same profile.

\section{{\sffamily\scshape Channels4World} Corpus}
\label{sec:corpus}

{\sffamily\scshape Channels4World} contains 100,000 procedurally generated outdoor scenes: 50,000 Urban, 25,000 Metropolitan, and 25,000 Suburban.
Each scene aligns one environment and a shared BS/UE placement set with multipath evaluations at 3.5 and 7~GHz.
This section characterizes the resulting environment and placement distributions, summarizes the retained channel records, validates corpus integrity and the canonical public split, and describes the release organization and scope.

With 32 BS and 16,384 UE positions per scene, the corpus represents 52.43 billion physical BS--UE pairs and 104.86 billion frequency-specific evaluations.
Of these evaluations, 81.37\% contain at least one retained path, yielding 85.32 billion retained links, 800.89 billion retained paths, and approximately 1.47 trillion interaction records.
The packaged corpus payload totals 29.14~TB (26.50~TiB).

\subsection{Environment and Placement Characteristics}

We first characterize the variation realized by the three generation profiles in environment morphology, material composition, and transceiver placement.
Fig.~\ref{fig:environment_placement_characterization} summarizes these distributions and also compares three-dimensional BS--UE range to assess whether the profile differences simply reflect different link-distance distributions.

Material shares are computed by building count, and built fraction is the fraction of scene area covered by building footprints.
Panels~(f) and~(g) use a deterministic sample of 64 UEs per scene paired with all 32 BSs, yielding 2,048 BS--UE pairs per scene for the UE-clearance and range analyses.

Building density, built fraction, footprint size, height, material composition, BS elevation, and UE clearance vary within and across the three profiles.
The profile distributions are distinct but overlapping.
Metropolitan scenes occupy the high-density, high-built-fraction region and generally contain taller buildings.
Suburban scenes tend to contain fewer buildings with larger footprints, while Urban scenes span an intermediate region.
The realized material shares closely follow the profile-specific sampling weights.

\begin{table*}[!t]
\caption{Full-corpus retained-channel summary by profile and frequency.}
\label{tab:profile_carrier_census}
\centering
\footnotesize
\setlength{\tabcolsep}{8pt}
\renewcommand{\arraystretch}{1.10}
\begin{tabularx}{\textwidth}{@{}L{2.6cm}L{1.8cm}L{2.5cm}L{2.1cm}L{2.5cm}Y@{}}
\toprule
\textbf{Profile} & \textbf{Frequency} & \textbf{Retained links (bil.)} & \textbf{Retention (\%)} & \textbf{Retained paths (bil.)} & \textbf{Paths / retained link} \\
\midrule
Urban & 3.5~GHz & 23.35 & 89.07 & 257.78 & 11.04 \\
Urban & 7~GHz & 21.13 & 80.61 & 188.93 & 8.94 \\
Metropolitan & 3.5~GHz & 9.02 & 68.84 & 87.81 & 9.73 \\
Metropolitan & 7~GHz & 7.42 & 56.58 & 62.75 & 8.46 \\
Suburban & 3.5~GHz & 12.40 & 94.63 & 119.70 & 9.65 \\
Suburban & 7~GHz & 12.00 & 91.52 & 83.91 & 7.00 \\
\midrule
All profiles & 3.5~GHz & 44.78 & 85.40 & 465.29 & 10.39 \\
All profiles & 7~GHz & 40.54 & 77.33 & 335.60 & 8.28 \\
All profiles & Aggregate & 85.32 & 81.37 & 800.89 & 9.39 \\
\bottomrule
\end{tabularx}
\end{table*}

The placement panels show generally higher BS elevations in Metropolitan scenes and greater UE clearance from buildings in Suburban scenes.
Despite these environmental and placement differences, the three-dimensional BS--UE range distributions remain broadly similar across profiles.
The profile distinctions therefore do not simply reflect different link-distance distributions.

\subsection{Retained Channel Summary}

Table~\ref{tab:profile_carrier_census} summarizes the full-corpus retained-link and retained-path outcomes.
Here, retention is the fraction of frequency-specific evaluations containing at least one retained path.
Because the Urban profile contains twice as many scenes as either Metropolitan or Suburban, the raw retained-link and retained-path counts are influenced by corpus composition.
Retention and paths per retained link provide the more direct cross-profile comparisons.

Overall retention decreases from 85.40\% at 3.5~GHz to 77.33\% at 7~GHz, while the mean number of retained paths per retained link falls from 10.39 to 8.28.
The same frequency trend appears in every profile: 7-GHz evaluations are less often retained, and the retained links contain fewer stored paths.
Suburban scenes have the highest retention at both frequencies, while Metropolitan scenes have the lowest.
Together with the broadly similar BS--UE range distributions in Fig.~\ref{fig:environment_placement_characterization}, this profile ordering is not simply a consequence of different link-distance distributions.
Section~\ref{sec:evaluation} examines the retained multipath behavior in greater detail.

\subsection{Corpus Integrity and Scene-Level Split}
\label{subsec:record_integrity}

We validate the release at three levels: sampled regeneration of scene assets, consistency of the serialized channel representation, and integrity of scene identity and the public train--test split.

During qualification, replay checks regenerated the geometry and placements for six sampled scenes and matched the stored artifact hashes exactly.
This replay verifies deterministic regeneration of the sampled environment and placement assets; it does not establish identical regeneration of the channel and interaction records.
Separately, a synthetic-fixture validator reconstructed CIRs and CFRs from serialized path records at both carriers, with all comparisons satisfying the prescribed numerical tolerances.
The fixture definitions, delay references, frequency offsets, tolerances, and observed reconstruction errors are detailed in Appendix~\ref{app:record_validation}.
These checks establish consistency of the stored channel representation, but not real-world propagation accuracy.

Exact-identity checks confirm that all 100,000 scene records are distinct under geometry-only, geometry-plus-material, placement, and complete-record comparisons.
These checks establish exact distinctness of the serialized scenes, but do not quantify how geometrically similar two different scenes may be.

The canonical public split is defined at the scene level because links within the same scene share an underlying environment and are spatially correlated.
It contains 90,000 training and 10,000 test scenes, with the corpus's 50:25:25 Urban, Metropolitan, and Suburban composition preserved exactly in both.
Geometry, material assignments, BS and UE positions, channel records at both frequencies, and ordered interaction records remain together for every scene.
Because split membership is attached to scene identity, it is invariant across frequencies and corpus components, and no scene contributes records to both the training and test sets.
No separate canonical validation split is defined for the public release.

Together, these checks establish sampled-asset replay, stored-channel reconstruction agreement, exact scene distinctness, and the absence of same-scene leakage.

\subsection{Component-Wise Access and Nested Scene Subsets}

{\sffamily\scshape Channels4World} is released publicly alongside this article through the project website at \url{https://www.cbchae.org/}.
The release design indexes each scene by split, generation profile, frequency, and corpus component.
Geometry and placements are stored separately from frequency-specific channel and ordered interaction records, allowing workflows to retrieve only the required components instead of transferring the complete packaged corpus.

The release design defines deterministic nested scene subsets containing 200, 1,000, 10,000, 50,000, and 100,000 scenes.
Smaller subsets reduce storage, transfer, and compute requirements, while nesting allows experiments to expand in scale without replacing scenes already included.
All subsets preserve scene identity, inherit the canonical train--test assignments, and retain the overall generation-profile composition.

\begin{figure*}[t]
\centering
\includegraphics[width=\textwidth]{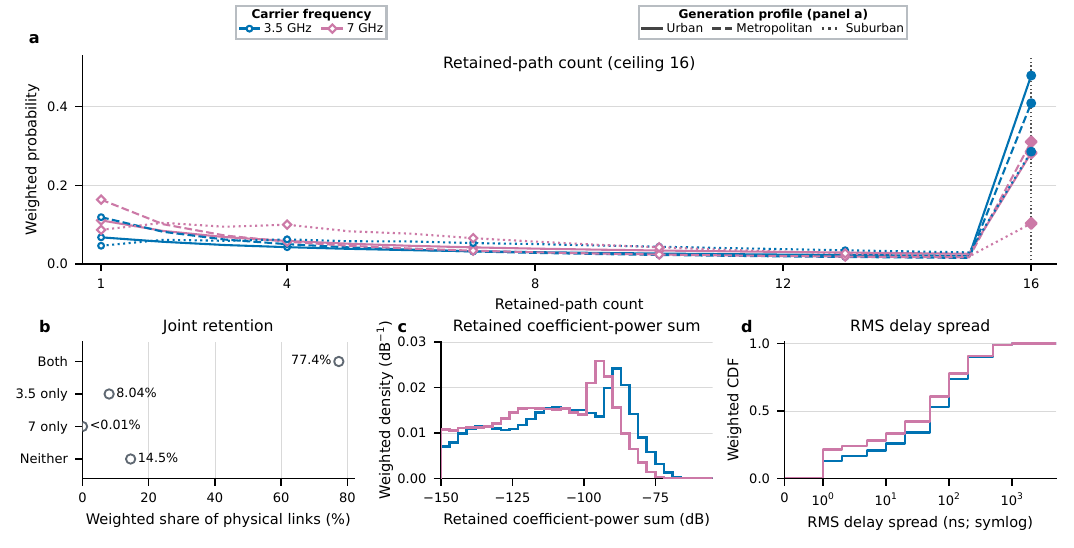}
\caption{Design-weighted retained-channel characteristics from the 3,000-scene morphology-stratified sample.
Carrier frequency is distinguished by color, while panel (a) additionally uses line style for generation profile.
Panel (b) uses aligned physical BS--UE pairs, while panels (a), (c), and (d) are conditioned on retained links.}
\label{fig:retained_multipath_characterization}
\end{figure*}

\subsection{Dataset Scope and Limitations}
\label{subsec:scope_boundaries}

{\sffamily\scshape Channels4World} represents static, procedural outdoor environments under three configured generation profiles.
Each environment is limited to a \(500\times500\)~m domain, so the channel records do not include paths involving structures beyond its boundary.
Buildings use simplified exterior envelopes, and the current scenes omit terrain elevation, vegetation, facade apertures, street furniture, and indoor geometry.
They also omit weather, traffic, and moving blockers.
UE positions are independently sampled rather than organized into mobility trajectories.

The corpus fixes the material model, carriers, propagation mechanisms, and path-retention rule, while building-level material assignments vary according to the configured profile distributions.
For each retained link, the stored multipath record contains at most the 16 strongest candidate paths that also pass the strict power threshold, rather than every candidate returned by propagation simulation.
The path-level format supports downstream selection of compatible arrays, codebooks, and observation models.
The stored scalar coefficients use a V-polarized convention.

The procedural scenes are not reconstructions of deployed sites and have no one-to-one outdoor measurement counterparts.
Within the generated domain, held-out scenes contain separately generated realizations of roads, blocks, building morphology, material assignments, and transceiver placements.
Scene-disjoint evaluation therefore measures generalization across these procedural realizations.
Transfer to geographically grounded or measured environments remains a separate empirical question involving environmental shift, simulator error, observation choices, and hardware effects.

\section{Channel Analysis and Controlled Evaluation}
\label{sec:evaluation}

Having established the corpus characteristics, integrity, and scope, we next examine the retained channels in greater detail and present two controlled studies.
We first characterize cross-frequency multipath behavior and test how diffraction and path-retention settings affect the stored multipath representation and its storage requirements.
We then study how environmental breadth affects generalization to held-out scenes, with additional controls examining how training support is distributed across BS locations and how prediction depends on environment-specific context.

\begin{figure*}[t]
\centering
\includegraphics[width=\textwidth]{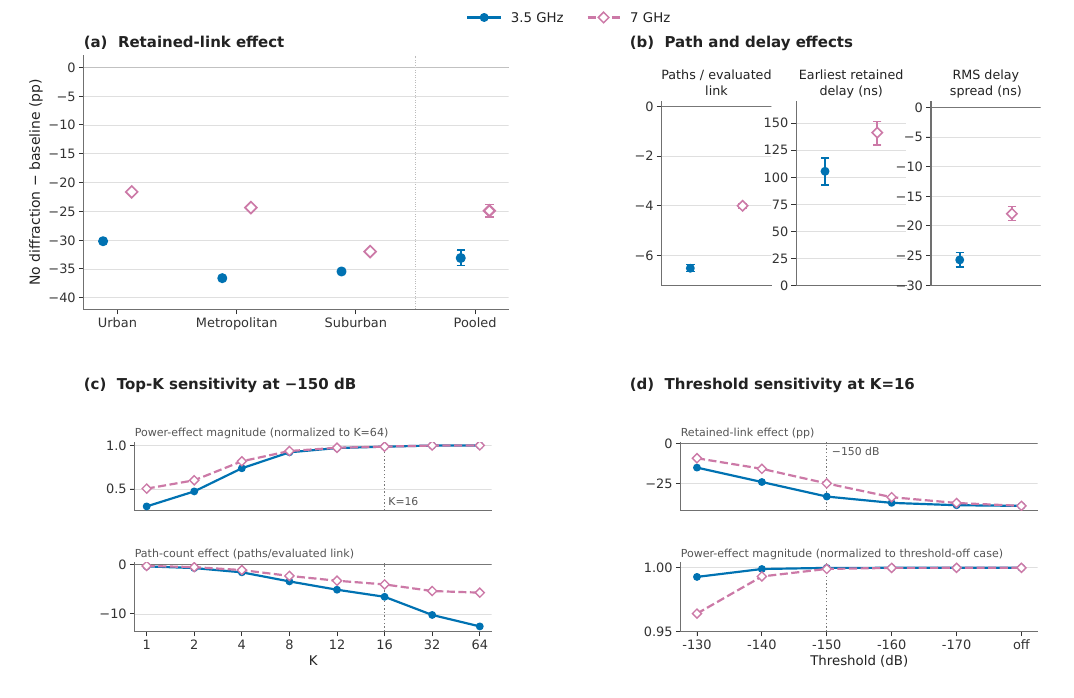}
\caption{Controlled sensitivity of the retained channel records to diffraction and path-retention settings across 48 scenes.
Direct effects are no diffraction minus baseline.
Panels (a)--(b) use the corpus \(K=16\), strict \(>-150\)-dB policy and show scene-stratified 95\% bootstrap intervals for pooled effects; panels (c)--(d) vary the path-count limit and power threshold.}
\label{fig:path_retention_sensitivity}
\end{figure*}

\subsection{Retained Multipath Characteristics Across Frequencies}

We first examine how the retained channel representation changes between 3.5 and 7~GHz when the physical environment and BS--UE placements are shared across carriers.
Beyond the full-corpus link and path counts in Table~\ref{tab:profile_carrier_census}, Fig.~\ref{fig:retained_multipath_characterization} compares path multiplicity, joint retention of aligned physical BS--UE pairs, retained coefficient-power sums, and delay spread.

The analysis uses the fixed 3,000-scene morphology-stratified sample and its inverse-selection design weights, described in Appendix~\ref{app:morphology_sampling}.
Panel~(a) resolves retained-path counts by profile and frequency, while panel~(b) compares retention outcomes for the same physical BS--UE pairs at both carriers.
Panels~(c) and~(d) show the retained coefficient-power sum and RMS delay spread, respectively.
Both are conditioned on retained links and aggregate across profiles using the corpus's 50:25:25 composition.

The retained 7-GHz records are generally sparser than their 3.5-GHz counterparts.
The top-16 ceiling is reached by 41.3\% of retained links at 3.5~GHz but by 24.5\% at 7~GHz.
This ceiling reflects truncation by the corpus retention policy rather than a physical upper bound on multipath richness, so evaluations at \(K_q=16\) may contain additional weaker candidate paths.

The aligned-pair comparison reveals a pronounced asymmetry in retention across frequencies.
Across the weighted sample, 77.4\% of physical BS--UE pairs are retained at both frequencies, 8.04\% only at 3.5~GHz, fewer than 0.01\% only at 7~GHz, and 14.5\% at neither.
Among pairs retained at both carriers, the 7-GHz records contain 2.97 fewer paths per link and have a retained coefficient-power sum 7.54~dB lower on average.
Their retained delay structure also differs: RMS delay spread is 14.1~ns lower, while the earliest retained path is 14.3~ns later.

These differences reflect both frequency-dependent propagation and the common retention policy.
For reference, doubling the carrier frequency from 3.5 to 7~GHz adds 6.02~dB of free-space path loss at fixed distance, while the observed retained-channel differences also depend on multipath propagation and which paths remain above the corpus retention threshold.

\subsection{Propagation and Path-Retention Sensitivity}
\label{subsec:propagation_path_selection_sensitivity}

\begin{table*}[!t]
\caption{Path-dependent storage for the 48-scene path-retention sweep.}
\label{tab:path_retention_storage}
\centering
\footnotesize
\setlength{\tabcolsep}{3pt}
\renewcommand{\arraystretch}{1.08}
\begin{tabularx}{0.72\textwidth}{@{}Yrr@{}}
\toprule
\textbf{Policy} & \textbf{Logical storage} & \textbf{vs. corpus policy}\\
\midrule
\(K=16,\ >-150\) dB (corpus policy) & 51.46 GB & 1.00\(\times\)\\
\(K=64,\ >-150\) dB & 78.15 GB & 1.52\(\times\)\\
\(K=16\), threshold disabled & 81.42 GB & 1.58\(\times\)\\
\(K=64\), threshold disabled & 206.40 GB & 4.01\(\times\)\\
\bottomrule
\end{tabularx}
\end{table*}

We next separate two sources of variation in the released records: the propagation mechanisms included during ray tracing and the path-retention policy applied afterward.
The study uses 48 fixed scenes (24 Urban, 12 Metropolitan, and 12 Suburban) selected from the canonical training split by profile-local hash ranking, without using morphology or channel outcomes.
Environment geometry and BS/UE placements are held fixed throughout.

For each scene, we compare the baseline corpus propagation setting with a variant in which diffraction is disabled.
Each setting is simulated twice, and the reported scene-level effects are averaged over the two paired repetitions.
Differences between repetitions are used to check repeatability; the repetitions are not counted as independent scenes.
The simulations retain the 64 strongest solver-returned paths before thresholding, allowing the top-\(K\) limit and power threshold to be varied afterward without rerunning the ray tracer.
These top-64 sets are specific to each simulation and do not represent all rays considered internally during path search.
In Fig.~\ref{fig:path_retention_sensitivity}, direct effects are defined as no diffraction minus baseline.

Removing diffraction produces substantial changes in both retained-link availability and multipath structure.
Under the corpus retention policy, the retained-link rate falls by 33.1 percentage points at 3.5~GHz and 24.9 percentage points at 7~GHz, while retained paths per evaluated link decrease by 6.52 and 4.00, respectively.
The delay structure also shifts: the earliest retained path is 105--141~ns later and RMS delay spread is 17.9--25.7~ns lower across the two carriers.
The pooled 95\% bootstrap intervals in Fig.~\ref{fig:path_retention_sensitivity} do not cross zero for these effects.
Aggregate power represented by the retained paths also decreases, although the profile-level power changes are less consistent.

\begin{figure*}[t]
\centering
\includegraphics[width=\textwidth]{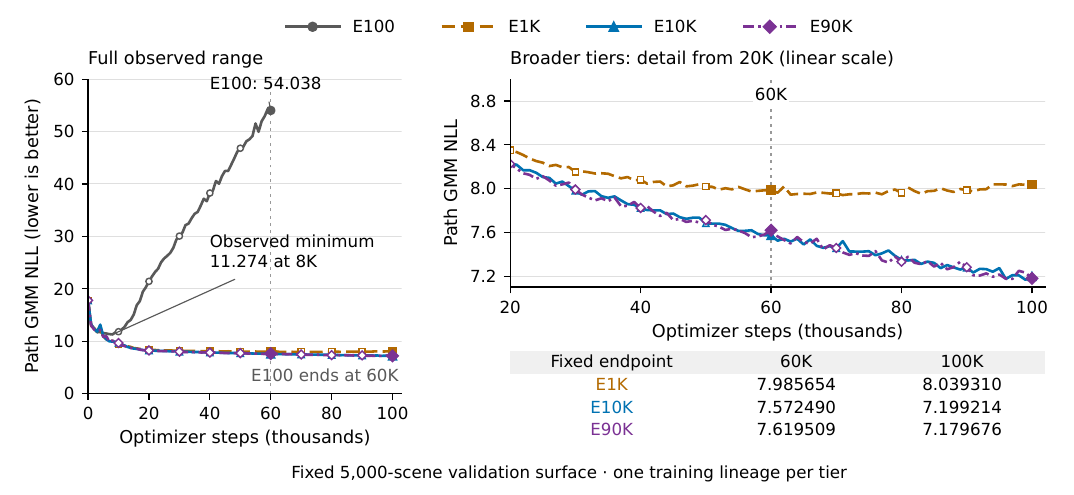}
\caption{Path GMM NLL validation trajectories for the environmental-breadth study at 7~GHz.
The common four-tier comparison is at 60K; E1K, E10K, and E90K are continued to 100K.}
\label{fig:environmental_scaling}
\end{figure*}

We then ask how much represented power is lost by limiting each retained link to its strongest 16 paths.
At the common strict \(>-150\)-dB threshold, \(K=16\) preserves 99.982\% and 99.989\% of the aggregate power represented by \(K=64\) at 3.5 and 7~GHz, respectively.
The strongest 16 paths therefore contain nearly all of the represented power in this study.
This does not mean that the remaining paths are irrelevant to every analysis: Fig.~\ref{fig:path_retention_sensitivity}(c) shows that the diffraction-induced path-count difference continues to grow beyond \(K=16\), even though \(K=16\) already captures about 98.6\% of the corresponding power change observed at \(K=64\).
Because the sweep stores only the 64 strongest solver-returned paths, the aggregate contribution of any weaker returned paths beyond that set is not measured here.

The power threshold has a different effect: it primarily determines whether evaluations containing only very weak candidate paths remain as retained links.
Relative to \(K=64\) with the threshold disabled, the corpus policy preserves more than 99.98\% of the represented aggregate power at both carriers while retaining 93.63\% of the reference links at 3.5~GHz and 84.66\% at 7~GHz.
The more permissive setting therefore adds retained links, especially at 7~GHz, but contributes very little additional represented power.

These retention choices also reduce the size of the path-dependent records substantially.
Here, logical path-dependent storage is the exact uncompressed size of the retained-link, path, and interaction arrays.
It excludes geometry, placements, compression, catalogs, and container or filesystem overhead.
As summarized in Table~\ref{tab:path_retention_storage}, storage across the 48 scenes decreases from 206.40~GB for \(K=64\) with the threshold disabled to 51.46~GB under the corpus policy, a 75.1\% reduction.

Applying the measured channel-core and interaction-detail ratios separately to the corresponding full-corpus layers projects a 115.05-TB packaged payload for the \(K=64\), threshold-disabled setting, compared with the measured 29.14-TB payload of the released corpus.
This projection holds the non-path scene layers fixed and does not represent a materialized or recompressed alternative corpus.

In this study, the retention policy substantially reduces storage while preserving nearly all of the aggregate power represented by the top-64 reference sets.
Analyses that depend on very weak links or detailed path multiplicity should account for the paths omitted by the retention policy.

\subsection{Environmental Breadth and Training Horizon}
\label{subsec:environmental_scaling}

The second controlled study asks how the breadth of training environments affects the learning of reusable RF-environment representations.
We adopt the pretraining formulation of Channel2World~\cite{moon2026channel2world}, which uses many wireless observations collected from one BS to infer a fixed-size ``world embedding'' representing the surrounding RF environment.
Its context--query formulation is particularly suitable for studying environmental breadth because the embedding is encouraged to capture propagation information shared across links within an environment.

The task follows an observe-then-predict structure.
For a given BS, the model first observes a set of \emph{context links} for which both multipath channels and UE positions are known and compresses them into the world embedding.
It then receives disjoint \emph{query links} from the same BS.
For each query path, only the BS-side path direction and propagation delay are provided, and the model predicts the UE location.

For a reflected path, direction and delay do not uniquely determine the UE location: they constrain how the path leaves the BS and its total length, but not where reflection occurs.
The set of plausible UE locations therefore depends on the surrounding RF environment.
The world embedding supplies this environment-specific information to the query predictor, encouraging it to retain propagation structure shared across links in the same environment.
Channel2World further showed that these embeddings can reconstruct dominant reflector structures around the BS~\cite{moon2026channel2world}.

We apply this formulation separately to individual BSs within each {\sffamily\scshape Channels4World} scene.
Each eligible scene--BS pair forms a \emph{BS-centered episode}, so one scene can contribute up to 32 episodes associated with different BS locations but the same underlying environment.
During training, a scene is sampled first and an eligible BS is then sampled from that scene.
Increasing the number of training scenes therefore increases environmental breadth while also increasing the pool of distinct BS-centered episodes available for sampling.

Each episode contains 192 context links and 32 disjoint query links, with up to 16 retained paths per link.
Because Channel2World uses uplink channels whereas {\sffamily\scshape Channels4World} stores downlink channels, we use BS-side departure angles in place of BS-side arrival angles.
The study uses the 7-GHz carrier throughout.
To better match the original Channel2World propagation configuration, retained paths involving diffraction are excluded from the learning inputs.
This filtering affects only the learning view; diffraction remains part of the released corpus.
Query links contain a retained LoS path, while all paths remaining after this learning-view filter, including reflected paths, are used for training and evaluation.
Because the model receives no scene geometry or RF material assignments directly, any environment-specific information available to the predictor must be inferred from the wireless context.

We use three complementary comparisons.
First, nested training tiers test how held-out-scene performance changes as training support spans more environments under a common exposure budget.
Second, an equal-episode control compares different allocations of the same number of available BS-centered episodes across environments and BS locations.
Third, inference-time controls test whether prediction depends on the identity and amount of correctly aligned environment context.

For the main comparison, we train separate models on four nested subsets of the canonical 90,000-scene training split: E100, E1K, E10K, and E90K, containing 100, 1,000, 10,000, and 90,000 scenes.
Each tier preserves the corpus's 50:25:25 Urban, Metropolitan, and Suburban composition.
The tiers are deterministic prefixes of one fixed profile-stratified training order, so every narrower tier is contained exactly in each broader tier.
Model architecture, objective, optimization settings, effective batch size, and training seed are fixed across tiers.
All reported comparisons use a single training seed and therefore do not quantify variability across independent training runs.

At the common 60K-step budget, every model processes 11.52 million BS-centered episodes sampled with replacement.
Narrower tiers consequently revisit the same environments more often, whereas broader tiers distribute the same exposure across more scenes.
The average number of sampled episodes per scene ranges from 115,200 for E100 to 128 for E90K.

For this study, we use a fixed 5,000-scene subset of the canonical 10,000-scene test split for validation and leave the remaining 5,000 scenes unused in the reported experiments.
Performance is measured by the negative log-likelihood (NLL) of the path-level Gaussian mixture model (GMM) at the true UE position, denoted Path GMM NLL.
Lower values indicate better prediction, and validation scenes are weighted equally.
The complete learning protocol and equal-episode-control settings are summarized in Table~\ref{tab:breadth_learning_protocol} in Appendix~\ref{app:training_details}, which also gives additional protocol and metric details.

Figure~\ref{fig:environmental_scaling} summarizes the validation trajectories for the four main training tiers.
The main training ladder shows lower validation NLL for the broader E10K and E90K tiers than for E1K at the common 60K endpoint, but the comparison also depends on training horizon.
At 60K, Path GMM NLL is 54.038 for E100, 7.986 for E1K, 7.572 for E10K, and 7.620 for E90K.
Both E10K and E90K therefore have lower NLL than E1K at the matched-exposure endpoint, while E100 has already deteriorated substantially after reaching its lowest observed NLL of 11.274 at 8K.

The trajectories show that 60K is not a common convergence point.
E1K changes little near the end of the 60K run, whereas E10K and E90K continue to improve, and E90K reaches 7.556 at 57K before ending at 7.620.
We therefore continue E1K, E10K, and E90K to 100K steps without restarting training; E100 is not extended because its validation performance has already deteriorated sharply.

At 100K, Path GMM NLL is 8.039 for E1K, 7.199 for E10K, and 7.180 for E90K.
E10K and E90K continue to improve beyond 60K, whereas E1K does not.
E10K and E90K differ only slightly, and their ordering reverses between the 60K and 100K endpoints.

\begin{figure*}[t]
\centering
\includegraphics[width=\textwidth]{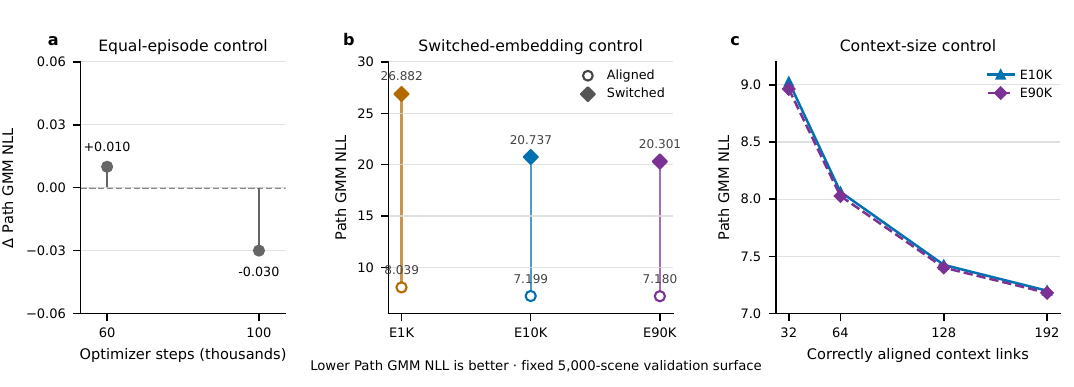}
\caption{Controls for the environmental-breadth study.
(a) Equal-episode comparison between E32K-BS1 and E1K.
(b) Correctly aligned versus same-profile switched environment embeddings.
(c) Inference-time sensitivity to the number of correctly aligned context links.}
\label{fig:environmental_controls}
\end{figure*}

Because broader tiers also contain more distinct BS-centered episodes, the main tiers do not vary environmental breadth in isolation.
We therefore add an equal-episode control with 32,000 available episodes in each condition: E1K uses 32 eligible episodes from each of 1,000 scenes, whereas E32K-BS1 uses one fixed eligible episode from each of 32,000 scenes.
This holds the available episode count fixed while shifting support from many BS locations within fewer environments to one BS location across many more environments.

Under matched evaluation, E32K-BS1 differs from E1K by only \(+0.010\) Path GMM NLL at 60K and \(-0.030\) at 100K, as shown in Fig.~\ref{fig:environmental_controls}(a).
The direction of the difference therefore changes with training horizon and its magnitude remains small.
This control shows that holding the available episode count fixed while distributing that support across many more scenes does not reproduce the substantially larger differences observed in the main training ladder.
It does not isolate scene count as a single causal variable, because the two conditions also differ in how BS locations are distributed within each environment.

We next test whether the world embedding carries environment-specific information that the predictor actually uses.
For the 100K E1K, E10K, and E90K checkpoints, each correctly aligned embedding is replaced by one inferred from a different validation scene of the same generation profile, while the recipient queries, targets, and model remain unchanged.
As shown in Fig.~\ref{fig:environmental_controls}(b), Path GMM NLL increases from 8.039 to 26.882 for E1K, from 7.199 to 20.737 for E10K, and from 7.180 to 20.301 for E90K.
The switched embedding performs worse on all 5,000 validation scenes for every tier.
The predictor therefore relies on environment-specific information in the correctly matched embedding rather than treating same-profile embeddings as interchangeable.

We also vary how much correctly aligned context is available at inference.
For E10K and E90K, fixed nested subsets of 32, 64, and 128 links are re-encoded from the original 192-link context while the trained model and query set remain unchanged.
Figure~\ref{fig:environmental_controls}(c) shows a monotonic decrease in Path GMM NLL as context increases from 32 to 192 links: from 9.027 to 7.199 for E10K and from 8.963 to 7.180 for E90K, with the intermediate 64- and 128-link conditions following the same trend.
Additional correctly aligned context therefore improves prediction throughout the tested range.
Because all models were trained with 192 context links, this is an inference-time context-sensitivity result rather than a comparison of models trained at different context sizes.

Taken together, these experiments separate several factors that are coupled in the main environmental-breadth comparison.
Broader E10K and E90K training support yields lower held-out-scene NLL than E1K under the tested training protocol, while the trajectories show that the relative behavior of the broader tiers also depends on training horizon.
When the available episode count is fixed, distributing that support across many more environments produces only a small, horizon-dependent difference, so the larger separations in the main ladder should not be attributed to scene count alone.
The inference-time controls further show that prediction depends strongly on both the identity and amount of correctly aligned environment context.
These results motivate future experiments that vary environmental breadth, the number and distribution of BS locations within each environment, model capacity, and training compute independently.

\section{Applications and Future Directions}
\label{sec:discussion}

{\sffamily\scshape Channels4World}'s scene-centered representation supports a range of downstream wireless tasks and modular corpus extensions.
Before outlining these uses and extensions, we summarize the limitations that bound the present corpus and controlled evaluations.

\subsection{Limitations of the Present Evaluation}
\label{subsec:discussion_limitations}

The corpus-level scope and representation boundaries are summarized in Section~\ref{subsec:scope_boundaries}.
Accordingly, the retained-channel characterization and propagation/path-retention sensitivity analyses in Section~\ref{sec:evaluation} apply to the released retained-path representation and controlled simulation variants.
These analyses do not establish real-world propagation accuracy or transfer to geographically grounded environments.

The learning study in Section~\ref{subsec:environmental_scaling} is descriptive rather than a definitive scaling study.
All reported training comparisons use a single training seed, one Channel2World-based architecture and objective, a fixed 7-GHz learning view, and the common optimization protocol summarized in Table~\ref{tab:breadth_learning_protocol}.
The main training ladder also increases the number of available BS-centered episodes as environmental breadth increases, while the equal-episode control changes how a fixed episode count is distributed across environments and BS locations.
The experiments therefore do not isolate scene count as a single causal variable.
The reported comparisons do not quantify variability across independent training runs or establish architecture-independent scaling laws or convergence behavior.

\subsection{Applications Across Diverse Environments}

The stored path parameters can be transformed into task-specific channel representations, including CIRs, CFRs, CSI tensors, array responses, and beam labels, once the required sampling, array, and codebook conventions are specified.
These representations can support channel prediction and synthesis, compression, representation learning, beam management, localization, and spatial inference.
Waveform, interference, noise, normalization, and hardware assumptions can be chosen for each task.
Position-indexed quantities derived from the stored UE positions and link records remain distinct from radio maps evaluated on a new receiver grid.

Aligned evaluation at 3.5 and 7~GHz supports cross-frequency prediction, transfer, and representation analysis while holding environment geometry, materials, and placements fixed.
The interaction summaries and ordered interaction records can further support multipath classification, physical-path analysis, and graph representations that connect transceivers and interaction points.
These path-centered representations can also be combined with geometry-aware models that operate on scene primitives~\cite{hehn2025wigatr}.
Together, the channel and interaction records, environment geometry, and material assignments support studies of how environmental structure relates to propagation behavior.

Existing channel-learning methods can be evaluated across large sets of held-out procedural scenes without changing their underlying task definitions.
Models can be trained across broad procedural variation and compared using the training partitions of the nested scene subsets or other controlled changes in environmental breadth.
This organization supports experimental designs that compare repeated exposure to fewer environments with training across broader environmental support while preserving scene-disjoint evaluation.
It also provides a common basis for studying representations intended for reuse across tasks or deployment settings~\cite{chen2026wirelessworldmodel,moon2026channel2world}.

\subsection{Extending the Corpus Through Modular Design}

The generation framework separates environment construction, transceiver placement, propagation simulation, and export, allowing individual stages to be changed while preserving the same scene-centered organization.
Extensions can therefore introduce new environment sources, deployment patterns, propagation settings, or derived observation layers without changing how scenes and their derived records are indexed and compared.
To preserve reproducibility, such extensions should record their source data and modified stages, together with parent-scene identity when a new result is derived from an existing scene.

Geographically grounded environments are a natural extension of the current procedural scene source.
Imported building and terrain data can replace the procedural geometry stage.
Downstream stages can be reused where their assumptions remain appropriate.
In practice, geographic sources may require geometry repair and completion of missing attributes such as building heights or materials, together with explicit handling of terrain, boundaries, licensing, and provenance~\cite{huangfu2022waird,li2023geo2sigmap,liu2025openpathnet}.

Placement can be varied independently of environment source.
Alternative BS deployments, UE distributions, and ordered UE trajectories can reuse a fixed environment, while dynamic studies can extend the scene representation with time, motion, Doppler, and other temporal information~\cite{klautau2021raymobtime}.

The same scene can also be resimulated under changed physical or system assumptions while holding geometry and placements fixed.
Changing carrier frequency, antenna or polarization conventions, propagation mechanisms, material models, or path-retention settings creates matched variants for studying how those choices affect the resulting channels.
Individual paths need not correspond one-to-one across such variants, so comparisons should remain tied to the shared scene and simulation setting rather than assumed path identity.

Radio maps provide a complementary observation layer.
The existing environments, material assignments, and BS positions can be reused, but native maps require a new ray-tracing pass over a specified receiver grid and observation configuration~\cite{hoydis2023sionnart,aoudia2025sionnart}.
The 100,000 scenes, 32 BS positions, and two carriers define up to 6.4 million BS--frequency map layers before additional heights, arrays, beams, or propagation settings are introduced.
This scale could support radio-map prediction and reconstruction under scene-disjoint evaluation, controlled variation in environmental breadth, and aligned cross-frequency transfer~\cite{yapar2022radiomapseer,li2023geo2sigmap,li2026u6gradiomap}.

\subsection{Research Opportunities from Scale and Extension}

The environmental scale of {\sffamily\scshape Channels4World} enables studies of how training coverage relates to generalization across environments.
Future experiments can ask how much environmental breadth is useful, which forms of scene variation contribute most to transfer, and whether gains observed across procedural scenes persist under geographically grounded evaluation.
Procedural and geographic scene collections matched on propagation-relevant environmental descriptors could support controlled studies of procedural-to-geographic transfer.

Matched corpus extensions can also test transfer across physical settings and observation representations.
Representations learned from one carrier, propagation configuration, deployment pattern, or observation layer can be evaluated under controlled changes to those conditions.
For example, representations learned from position-indexed multipath records could be transferred or adapted to tasks involving ordered trajectories, radio maps, or graph-based representations associated with related scenes.
Preserving scene identity across such extensions makes it possible to relate changes in model behavior to controlled changes in the environment or observation process.

Independent measured datasets provide a separate test of transfer beyond the simulated distribution.
Pretrained models can be evaluated or adapted using quantities available in measured resources, such as received power, path loss, delay spread, or downstream task labels~\cite{li2023geo2sigmap,alkhateeb2023deepsense}.
Such experiments can test whether representations learned across simulated environments remain useful under geographic, propagation-model, and hardware differences.
However, agreement with measurements would not validate individual {\sffamily\scshape Channels4World} records.

Performance degradation or systematic errors across these evaluations can help identify where additional data would be most informative.
Errors can be grouped by environmental characteristics, material composition, source, or simulation setting and used to motivate targeted procedural generation, geographic acquisition, or matched resimulation.
Corpus expansion can then target observed coverage gaps rather than simply add more scenes.

In the longer term, the linked scene and channel records can support learning in both directions between environment and propagation.
Models may predict channels from environmental structure, infer environmental properties from wireless observations, or transfer across system configurations~\cite{chen2026wirelessworldmodel,moon2026channel2world}.
As new observation layers are added, shared scene identity provides a common basis for connecting these tasks rather than treating each extension as an unrelated dataset.

\section{Conclusion}
\label{sec:conclusion}

This paper presented {\sffamily\scshape Channels4World}, a public scene-centered wireless dataset of 100,000 outdoor scenes, each binding one procedurally generated RF environment to aligned BS/UE placements and multipath and interaction records at 3.5 and 7~GHz.
The release preserves scene identity across frequencies and corpus components while making environmental breadth and within-environment observation density separately controllable.
Component-wise access, deterministic nested subsets, and a scene-level train--test split support experiments across different storage, compute, and data-scale requirements without breaking this alignment.

Corpus characterization shows substantial variation in environment morphology, material composition, and transceiver placement across the three generation profiles.
A controlled propagation study further shows that diffraction materially affects retained-link availability and multipath structure.
In the same study, the corpus retention policy preserves more than 99.98\% of the aggregate power represented by the richer retained sets and reduces path-dependent storage by 75.1\%.
These analyses characterize what information the released representation preserves and the tradeoffs introduced by its propagation and retention choices.

The learning study compares held-out-scene performance across training sets of different environmental breadth.
Broader environmental support can improve held-out-scene generalization under the tested protocol, while a matched episode-count control shows that scene count alone does not explain the larger differences in the main training ladder.
Inference-time controls further show that prediction depends strongly on correctly matched environment-specific context.
Together, these results support treating environmental breadth as a distinct dimension of wireless pretraining data rather than equating dataset scale with observation count alone.

By providing large numbers of distinct environments together with dense, aligned wireless observations, {\sffamily\scshape Channels4World} is intended to support the development and evaluation of wireless foundation models and other methods that must generalize across deployment environments.
Its scene-centered structure also provides a common basis for controlled extensions to geographic environments, alternative propagation settings, trajectories, radio maps, and other observation layers.
Transfer to geographically grounded or measured environments remains a separate empirical question.

\appendix[Validation and Evaluation Details]
\label{app:evaluation_details}

This appendix provides validation and evaluation details for Sections~\ref{sec:corpus} and~\ref{sec:evaluation}.
It covers serialized-record reconstruction, morphology-stratified sampling and weighting, the learning representation and objective, and the training and control protocols.

\subsection{Serialized-Record Reconstruction Validation}
\label{app:record_validation}

To verify the serialized path-record semantics independently of corpus-scale channel outcomes, a separate validator reconstructed CIRs and CFRs from two deterministic synthetic SISO fixtures: a 10-m free-space LoS link and a 10-m LoS-plus-single-specular link.
Each fixture was evaluated at 3.5 and 7~GHz using both absolute and first-path-normalized delay references, with CFR comparisons at baseband frequency offsets of \(-0.5\), \(0\), and \(+0.5\)~MHz.

Across two fresh-process repetitions, all 24 CIR and 48 CFR comparisons met the acceptance tolerances of \(5\times10^{-3}|a_{q,k}|+10^{-12}\) for CIR and \(5\times10^{-3}\sum_k|a_{q,k}|+10^{-12}\) for CFR.
The maximum complex absolute and scale-normalized errors were \(2.29\times10^{-8}\) and \(3.40\times10^{-5}\), respectively.

\subsection{Morphology-Stratified Channel Analysis}
\label{app:morphology_sampling}

The detailed multipath characterization in Section~\ref{sec:evaluation} uses a fixed 3,000-scene sample containing 1,500 Urban, 750 Metropolitan, and 750 Suburban scenes.
To preserve coverage of the morphology realized within each profile, the full scene population is stratified using building density, built-footprint fraction, and median building height.
Each descriptor is divided into profile-specific quintile bins, and the three bin indices jointly define a morphology stratum, yielding at most 125 strata per profile.
Scenes are selected deterministically within each stratum by SHA-256 ranking of scene identity.

Because the resulting sampling rates differ across morphology strata, the channel statistics in Fig.~\ref{fig:retained_multipath_characterization} use inverse-selection design weights.
For a populated stratum \(h\), let \(N_h\) denote its full-corpus scene count and \(n_h\) its selected count.
Each sampled scene in that stratum receives the unnormalized weight
\[
w_h=\frac{N_h}{n_h}.
\]
For each reported statistic, the weights are renormalized after applying the relevant conditioning, such as retention at one carrier or at both carriers.
This weighting corrects the deliberate morphology stratification of the analysis sample; it does not assign real-world geographic prevalence to the procedural profiles.
Fig.~\ref{fig:retained_multipath_characterization} states the retained-link or physical-pair conditioning used by each panel.

\subsection{Learning Representation and Objective}
\label{app:learning_representation}

\subsubsection{Episode and Feature Representation}

The environmental-breadth study uses a filtered view of the stored 7-GHz records rather than a separate propagation simulation.
To match the propagation scope of Channel2World~\cite{moon2026channel2world}, diffraction-flagged retained paths are removed.
For each eligible scene--BS pair, deterministic hash ordering selects 32 query links containing a retained LoS path and 192 disjoint context links from the remaining nonempty links.
All paths remaining after this learning-view filter are used, including reflected paths on LoS-eligible query links.
Within each cached BS-centered episode, the context and query sets remain fixed.

Because {\sffamily\scshape Channels4World} contains downlink channels, the learning representation uses BS-side departure angles rather than the BS-side arrival angles used by the original uplink Channel2World formulation.
Let \(\mathbf{r}=\mathbf{x}^{\mathrm U}_{s,u}-\mathbf{x}^{\mathrm B}_{s,b}\) denote the UE position relative to the episode's BS, and let \(\theta_t\) and \(\phi_t\) denote a path's departure zenith and azimuth angles in the scene frame defined in Section~\ref{sec:dataset_design}.
With \(t\) denoting the transmit (BS) side, we encode these angles as \(\mathbf{a}_t=[\sin\theta_t,\cos\theta_t,\sin\phi_t,\cos\phi_t]^{\mathsf T}\).
The context and query feature vectors are
\begin{equation}
\begin{aligned}
\mathbf{f}^{\mathrm c}
&=\bigl[10^7(\tau-\tau_{\min}),\mathbf{a}_t^{\mathsf T},g_{\rm rel},\mathbf{r}^{\mathsf T}\bigr]^{\mathsf T},\\
\mathbf{f}^{\mathrm q}
&=\bigl[10^7\tau,\mathbf{a}_t^{\mathsf T}\bigr]^{\mathsf T}.
\end{aligned}
\label{eq:learning_features}
\end{equation}
Here, delays are in seconds, positions are in meters, and \(g_{\rm rel}=10\log_{10}(P/P_{\max})\) in dB with \(P=|a|^2\).
The quantities \(\tau_{\min}\) and \(P_{\max}\) are computed from the surviving paths of the same link.
Query position and relative gain are withheld from the query features and used as prediction targets.

Context and query features are standardized separately using statistics computed from each training tier.
During training, Gaussian perturbations are applied after standardization with standard deviations of 0.01 for delay, \(\pi/180\) for each angular feature, 1.5 for context gain, and 0.1 for each context-position coordinate.
These perturbations operate in normalized feature coordinates and serve as training augmentation rather than a physical measurement-error model.

\subsubsection{Model and Objective}

The model contains 83,501,266 trainable parameters and follows the architecture summarized in Table~\ref{tab:learning_architecture}.
Learned latent tokens aggregate the context-path features into an environment representation that interacts with each query path.
The path-position Gaussian mixture model (GMM) is predicted before within-link query pooling, whereas the channel-position GMM uses the pooled representation.
The post-attention path features also support relative-gain prediction.

\begin{table}[t]
\caption{Architecture used in the learning study.}
\label{tab:learning_architecture}
\centering
\footnotesize
\setlength{\tabcolsep}{3pt}
\renewcommand{\arraystretch}{1.08}
\begin{tabularx}{\columnwidth}{@{}L{0.43\columnwidth}Y@{}}
\toprule
\textbf{Block} & \textbf{Configuration}\\
\midrule
Shared attention / feed-forward & Width 384; 6 heads; feed-forward width 1536\\
Context within-link attention & 4 self-attention layers\\
Environment aggregation & 32 latent tokens; 6 latent cross/self-attention pairs; 5 reciprocal context cross/self-attention pairs\\
Query--environment interaction & 8 bidirectional cross-attention stages\\
Query within-link post-attention & 2 self-attention layers\\
Prediction heads & Two 768-wide GELU hidden layers; separate path-mixture logit, mean, and scale heads\\
Position distributions & Path- and channel-level GMMs with 8 spherical Gaussian components each\\
\bottomrule
\end{tabularx}
\end{table}

For the position heads, let \(S_{\rm pos}=100\) and define the normalized position \(\mathbf{z}=\mathbf{r}/S_{\rm pos}\), with \(\mathbf r\) expressed numerically in meters.
Each head predicts
\begin{equation}
p_{\mathbf z}(\mathbf z)
=\sum_{k=1}^{8}\pi_k\,\mathcal{N}\!\left(\mathbf z;\boldsymbol{\mu}_k,\sigma_k^2\mathbf I_3\right),
\label{eq:learning_gmm}
\end{equation}
where \(\boldsymbol{\mu}_k\in\mathbb{R}^3\) is the normalized-coordinate mean of component \(k\), \(\mathbf I_3\) is the \(3\times3\) identity matrix, and \(\pi_k\) is obtained by softmax over the predicted logits.
The component scale is \(\sigma_k=\max\{0.005,\exp[\min(h_k,4)]\}\), where \(h_k\) is the predicted log-scale.
The lower bound corresponds to a physical standard deviation of 0.5~m.
All paths of a query link share the same UE-position target.

The model jointly optimizes path-position, relative-gain, channel-position, and component-coverage objectives.
For each logical training group \(g\),
\begin{equation}
\begin{aligned}
\mathcal{L}_g={}&\mathcal{L}_{{\rm path},g}
+0.05\mathcal{L}_{{\rm gain},g}\\
&+0.5\mathcal{L}_{{\rm channel},g}
+0.05\mathcal{L}_{{\rm cov},g}.
\end{aligned}
\label{eq:learning_loss}
\end{equation}
The path- and channel-position terms are normalized-coordinate GMM negative log-likelihoods averaged over valid query paths and query links, respectively.
The gain term uses Smooth-L1 loss with \(\beta=3\)~dB.
The coverage term applies a Smooth-L1 penalty to the distance from the target position to the nearest predicted mixture-component mean, with the distance normalized by 10~m.

\subsection{Training, Evaluation, and Controlled Comparisons}
\label{app:training_details}

\subsubsection{Training and Validation}

The common learning protocol and equal-episode-control settings are summarized in Table~\ref{tab:breadth_learning_protocol}.

\begin{table}[t]
\caption{Environmental-breadth learning protocol and equal-episode control.}
\label{tab:breadth_learning_protocol}
\centering
\footnotesize
\setlength{\tabcolsep}{3pt}
\renewcommand{\arraystretch}{1.08}
\begin{tabularx}{\columnwidth}{@{}L{0.30\columnwidth}Y@{}}
\toprule
\textbf{Property} & \textbf{Setting}\\
\midrule
Carrier / learning view & 7~GHz / learning view excludes diffraction-flagged retained paths\\
Main training tiers & 100 / 1K / 10K / 90K scenes\\
Equal-episode control & E1K: 1K scenes, 32 eligible BS episodes/scene; E32K-BS1: 32K scenes, one fixed BS episode/scene; 32K available episodes each\\
Episode & One eligible scene--BS pair; 192 context and 32 disjoint LoS-eligible query links; up to 16 retained paths/link\\
Model / output & 83.50M parameters; 32 environment tokens; eight-component position GMMs and relative-gain prediction\\
Effective global batch & 192 BS-centered episodes per optimizer step\\
Optimizer & AdamW~\cite{loshchilov2019decoupled}; base learning rate \(1.2\times10^{-4}\); weight decay 0.01; gradient clipping 1.0\\
Learning-rate schedule & 2K-step warmup; cosine decay to \(1.2\times10^{-5}\) at 20K; constant thereafter\\
Training horizon & 60K common endpoint for main tiers and equal-episode control; E1K, E10K, E90K, and E32K-BS1 continued to 100K\\
Training seed & 13\\
Validation & Fixed 5K held-out scenes; no augmentation; equal-scene aggregation\\
Validation / checkpoints & Every 1K / every 5K steps\\
\bottomrule
\end{tabularx}
\end{table}

Training scenes are ranked independently within each generation profile by SHA-256 under a fixed learning-order domain and then interleaved in the repeating Urban--Metropolitan--Urban--Suburban order.
E100, E1K, E10K, and E90K contain the first 100, 1,000, 10,000, and 90,000 scenes of this common order, respectively.
Scenes are sampled with replacement, so narrower tiers revisit the same environments more frequently under the fixed exposure budget.

Validation uses a fixed set of 5,000 held-out scenes (2,500 Urban, 1,250 Metropolitan, and 1,250 Suburban) without augmentation and with one fixed BS-centered episode per scene.
For scene \(s\), let \(\mathcal V_s\) denote its valid query paths.
The primary metric is
\[
\ell_s
=
-\frac{1}{|\mathcal V_s|}
\sum_{i\in\mathcal V_s}
\log p_{\mathbf z,i}(\mathbf r_i/S_{\rm pos})
+3\log S_{\rm pos}.
\]
Logarithms are natural.
The additive term converts NLL to meter coordinates because \(p_{\mathbf r}(\mathbf r)=p_{\mathbf z}(\mathbf r/S_{\rm pos})/S_{\rm pos}^3\).

Reported Path GMM NLL is the equal-scene mean of \(\ell_s\), so every validation scene contributes equally regardless of its number of valid query paths.
Validation is performed every 1K steps and checkpoints are saved every 5K steps, so the lowest observed validation point need not correspond to a saved checkpoint.

\subsubsection{Equal-Episode Control}

The equal-episode control holds the available episode count at 32,000 in both conditions while changing how those episodes are distributed across environments and BS locations.
E1K uses all 32 eligible BS-centered episodes from each of 1,000 scenes, whereas E32K-BS1 uses the first 32,000 scenes of the same common training order and one fixed eligible episode from each scene.
For E32K-BS1, the episode is selected deterministically from the eligible BS set and remains fixed throughout training.
At a given training horizon, the two conditions therefore match both available episode count and aggregate training exposure while distributing that support differently across environments and BS locations.
The control does not isolate scene count as a single causal variable.
The 60K and 100K endpoints are evaluated on the same fixed validation population using the same metric.

\subsubsection{Inference-Time Context Controls}

The switched-embedding control tests whether prediction depends on environment-specific information in the learned representation.
A fixed profile-preserving mapping assigns each validation episode a different donor episode from the same generation profile, with no self-mappings.
The donor context is encoded by the same trained model, and the resulting environment representation replaces the recipient representation while the recipient queries and targets remain unchanged.

The context-size control instead preserves the correct environment while varying the amount of available context.
For E10K and E90K, deterministic nested subsets of 32, 64, and 128 links are re-encoded from the original 192-link context while the trained model and query set remain unchanged.
The 192-link condition is the original full-context endpoint.
No model is retrained and no normalization statistics are recomputed, so the comparison measures inference-time sensitivity to context quantity rather than training with different context sizes.
Both inference-time controls use the same equal-scene aggregation over the fixed 5,000-scene validation population.

{\bibliographystyle{IEEEtran}}
\begingroup
\small
\balance
\bibliography{Channels4World_v1}

@string{ieee_m_comm         = {{IEEE} Commun. Mag.}}

@string{ieee_m_wireless     = {{IEEE} Wireless Commun.}}

@string{ieee_m_network      = {{IEEE} Netw.}}

@string{ieee_j_jsac         = {{IEEE} J. Sel. Areas Commun.}}

@string{ieee_j_wcom         = {{IEEE} Trans. Wireless Commun.}}

@string{ieee_j_comsurvtut   = {{IEEE} Commun. Surveys Tuts.}}

@string{ieee_j_ojcoms       = {{IEEE} Open J. Commun. Soc.}}

@string{ieee_j_tcomm        = {{IEEE} Trans. Commun.}}

@string{ieee_j_tap          = {{IEEE} Trans. Antennas Propag.}}

@string{ieee_j_tccn         = {{IEEE} Trans. Cogn. Commun. Netw.}}

@string{ieee_j_tnse         = {{IEEE} Trans. Netw. Sci. Eng.}}

@string{china_comm          = {China Commun.}}

@string{sci_j_scis          = {Sci. China Inf. Sci.}}

@string{conf_ita        = {Proc. Inf. Theory Appl. Workshop (ITA)}}

@string{conf_ssp        = {Proc. IEEE Stat. Signal Process. Workshop (SSP)}}

@string{conf_vtc_spring = {Proc. IEEE Veh. Technol. Conf. (VTC-Spring)}}

@string{conf_globecomwh = {Proc. IEEE Glob. Commun. Conf. (GLOBECOM) Workshops}}

@string{conf_globecom   = {Proc. IEEE Glob. Commun. Conf. (GLOBECOM)}}

@string{conf_icc        = {Proc. IEEE Int. Conf. Commun. (ICC)}}

@string{conf_dyspan     = {Proc. IEEE Int. Symp. Dyn. Spectr. Access Netw. (DySPAN)}}

@string{conf_iclr       = {Proc. Int. Conf. Learn. Represent. (ICLR)}}

@string{conf_icmlcn     = {Proc. IEEE Int. Conf. Mach. Learn. Commun. Netw. (ICMLCN)}}

@string{conf_icnc       = {Proc. Int. Conf. Comput. Netw. Commun. (ICNC)}}

@string{conf_milcom     = {Proc. IEEE Mil. Commun. Conf. (MILCOM)}}

@string{conf_iceti      = {Proc. Int. Conf. Emerg. Technol. Dependable Internet Things (ICETI)}}

@MISC{itu_imt2030,
  author       = {{International Telecommunication Union Radiocommunication Sector}},
  title        = {Framework and Overall Objectives of the Future Development of {IMT} for 2030 and Beyond},
  howpublished = {Recommendation ITU-R M.2160-0},
  year         = {2023}
}

@MISC{itu2023p2040,
  author       = {{International Telecommunication Union}},
  title        = {Effects of Building Materials and Structures on Radiowave Propagation Above About 100 {MHz}},
  howpublished = {Recommendation ITU-R P.2040-3},
  year         = {2023},
  month        = aug
}

@ARTICLE{liu2012cost2100,
  author  = {Liu, Lingfeng and others},
  title   = {The {COST 2100 MIMO} Channel Model},
  journal = ieee_m_wireless,
  year    = {2012},
  month   = dec,
  volume  = {19},
  number  = {6},
  pages   = {92--99},
  doi     = {10.1109/MWC.2012.6393523}
}

@TECHREPORT{3gpp38901,
  author      = {{3GPP}},
  title       = {Study on channel model for frequencies from 0.5 to 100 {GHz}},
  institution = {3rd Generation Partnership Project (3GPP)},
  number      = {TR 38.901},
  version     = {19.4.0},
  month       = jul,
  year        = {2026}
}

@ARTICLE{jaeckel2014quadriga,
  author  = {Jaeckel, Stephan and others},
  title   = {{QuaDRiGa}: A {3-D} Multi-Cell Channel Model with Time Evolution for Enabling Virtual Field Trials},
  journal = ieee_j_tap,
  year    = {2014},
  month   = jun,
  volume  = {62},
  number  = {6},
  pages   = {3242--3256},
  doi     = {10.1109/TAP.2014.2310220}
}

@INPROCEEDINGS{Klautau18,
  author    = {Klautau, Aldebaro and others},
  title     = {{5G} {MIMO} Data for Machine Learning: Application to Beam-Selection Using Deep Learning},
  booktitle = conf_ita,
  year      = {2018},
  month     = feb,
  pages     = {1--9},
  doi       = {10.1109/ITA.2018.8503086}
}

@INPROCEEDINGS{alkhateeb2019deepmimo,
  author    = {Alkhateeb, Ahmed},
  title     = {{DeepMIMO}: A Generic Deep Learning Dataset for Millimeter Wave and Massive {MIMO} Applications},
  booktitle = conf_ita,
  year      = {2019},
  month     = feb,
  pages     = {1--8}
}

@INPROCEEDINGS{alrabeiah2019viwi,
  author    = {Alrabeiah, Muhammad and others},
  title     = {{ViWi}: A Deep Learning Dataset Framework for Vision-Aided Wireless Communications},
  booktitle = conf_vtc_spring,
  year      = {2020},
  pages     = {1--5},
  doi       = {10.1109/VTC2020-Spring48590.2020.9128579}
}

@INPROCEEDINGS{klautau2021raymobtime,
  author    = {Klautau, Aldebaro and others},
  title     = {Generating {MIMO} Channels for {6G} Virtual Worlds Using Ray-Tracing Simulations},
  booktitle = conf_ssp,
  year      = {2021},
  month     = jul,
  pages     = {595--599},
  doi       = {10.1109/SSP49050.2021.9513861}
}

@MISC{huangfu2022waird,
  author        = {Huangfu, Yourui and others},
  title         = {{WAIR-D}: Wireless {AI} Research Dataset},
  howpublished  = {arXiv:2212.02159},
  year          = {2022},
  eprint        = {2212.02159},
  archivePrefix = {arXiv}
}

@INPROCEEDINGS{shen2023dataai6g,
  author    = {Shen, Zibing and others},
  title     = {{DataAI--6G}: A System Parameters Configurable Channel Dataset for {AI--6G} Research},
  booktitle = conf_globecomwh,
  year      = {2023},
  month     = dec,
  pages     = {1910--1915},
  doi       = {10.1109/GCWkshps58843.2023.10464657}
}

@ARTICLE{cheng2023m3sc,
  author  = {Cheng, Xiang and others},
  title   = {{M3SC}: A Generic Dataset for Mixed Multi-Modal Sensing and Communication Integration},
  journal = china_comm,
  year    = {2023},
  month   = nov,
  volume  = {20},
  number  = {11},
  pages   = {13--29},
  doi     = {10.23919/JCC.fa.2023-0268.202311}
}

@ARTICLE{alkhateeb2023deepsense,
  author  = {Alkhateeb, Ahmed and others},
  title   = {{DeepSense 6G}: A Large-Scale Real-World Multi-Modal Sensing and Communication Dataset},
  journal = ieee_m_comm,
  year    = {2023},
  month   = sep,
  volume  = {61},
  number  = {9},
  pages   = {122--128},
  doi     = {10.1109/MCOM.006.2200730}
}

@INPROCEEDINGS{shakya2025pointdata,
  author    = {Shakya, Dipankar and others},
  title     = {Standardized Machine-Readable Point-Data Format for Consolidating Wireless Propagation Across Environments, Frequencies, and Institutions},
  booktitle = conf_milcom,
  year      = {2025},
  month     = oct,
  pages     = {232--237},
  doi       = {10.1109/MILCOM64451.2025.11309979}
}

@INPROCEEDINGS{mao2026multimodalwireless,
  author    = {Mao, Tianhao and others},
  title     = {{Multimodal-Wireless}: A Large-Scale Dataset for Sensing and Communication},
  booktitle = conf_icc,
  month     = may,
  year      = {2026},
  pages     = {1--6},
  doi       = {10.1109/ICC59461.2026.11587897}
}

@MISC{deepmimo2026v4,
  author       = {{DeepMIMO Project}},
  title        = {{DeepMIMO} Documentation, Version 4},
  howpublished = {[Online]. Available: \url{https://www.deepmimo.net/docs/}},
  year         = {2026}
}

@ARTICLE{levie2021radiounet,
  author  = {Levie, Ron and others},
  title   = {{RadioUNet}: Fast Radio Map Estimation with Convolutional Neural Networks},
  journal = ieee_j_wcom,
  year    = {2021},
  month   = jun,
  volume  = {20},
  number  = {6},
  pages   = {4001--4015},
  doi     = {10.1109/TWC.2021.3054977}
}

@MISC{yapar2022radiomapseer,
  author        = {Yapar, {\c{C}}a{\u{g}}kan and others},
  title         = {Dataset of Pathloss and {ToA} Radio Maps With Localization Application},
  howpublished  = {arXiv:2212.11777},
  year          = {2022},
  eprint        = {2212.11777},
  archivePrefix = {arXiv}
}

@INPROCEEDINGS{li2023geo2sigmap,
  author    = {Li, Yiming and others},
  title     = {{Geo2SigMap}: High-Fidelity {RF} Signal Mapping Using Geographic Databases},
  booktitle = conf_dyspan,
  year      = {2024},
  month     = may,
  pages     = {277--285},
  doi       = {10.1109/DySPAN60163.2024.10632773}
}

@ARTICLE{zeng2024ckm,
  author  = {Zeng, Yong and others},
  title   = {A Tutorial on Environment-Aware Communications via Channel Knowledge Map for {6G}},
  journal = ieee_j_comsurvtut,
  year    = {2024},
  month   = {3rd Quart.},
  volume  = {26},
  number  = {3},
  pages   = {1478--1519},
  doi     = {10.1109/COMST.2024.3364508}
}

@ARTICLE{zhang2025spectrumnet,
  author  = {Zhang, Shuo and others},
  title   = {Generative {AI} on {SpectrumNet}: An Open Benchmark of Multiband {3-D} Radio Maps},
  journal = ieee_j_tccn,
  year    = {2025},
  volume  = {11},
  number  = {2},
  pages   = {886--901},
  doi     = {10.1109/TCCN.2024.3502492}
}

@ARTICLE{wu2025ckmimagenet,
  author  = {Wu, Zijian and others},
  title   = {{CKMImageNet}: A Dataset for {AI}-Based Channel Knowledge Map Toward Environment-Aware Communication and Sensing},
  journal = ieee_j_tcomm,
  year    = {2025},
  month   = dec,
  volume  = {73},
  number  = {12},
  pages   = {14430--14443},
  doi     = {10.1109/TCOMM.2025.3615778}
}

@ARTICLE{wang2025radiodiff3d,
  author  = {Wang, Xueyan and others},
  title   = {{RadioDiff-3D}: A {3D}$\times${3D} Radio Map Dataset and Generative Diffusion-Based Benchmark for {6G} Environment-Aware Communication},
  journal = ieee_j_tnse,
  year    = {2026},
  volume  = {13},
  pages   = {3773--3789},
  doi     = {10.1109/TNSE.2025.3590545}
}

@INPROCEEDINGS{jia2025urbanmimomap,
  author    = {Jia, Haoran and others},
  title     = {{UrbanMIMOMap}: A Ray-Traced {MIMO CSI} Dataset with Precoding-Aware Maps and Benchmarks},
  booktitle = conf_globecom,
  year      = {2025},
  month     = dec,
  pages     = {49--54},
  doi       = {10.1109/GLOBECOM59602.2025.11432757}
}

@MISC{li2026u6gradiomap,
  author        = {Li, Xiaojie and others},
  title         = {{U6G} {XL-MIMO} Radiomap Prediction: Multi-Config Dataset and Beam Map Approach},
  howpublished  = {arXiv:2603.06401},
  year          = {2026},
  eprint        = {2603.06401},
  archivePrefix = {arXiv}
}

@INPROCEEDINGS{hoydis2023sionnart,
  author    = {Hoydis, Jakob and others},
  title     = {{Sionna RT}: Differentiable Ray Tracing for Radio Propagation Modeling},
  booktitle = conf_globecomwh,
  year      = {2023},
  month     = dec,
  pages     = {317--321},
  doi       = {10.1109/GCWKSHPS58843.2023.10465179}
}

@MISC{aoudia2025sionnart,
  author        = {A{\"i}t Aoudia, Fay{\c{c}}al and others},
  title         = {{Sionna RT}: Technical Report},
  howpublished  = {arXiv:2504.21719},
  year          = {2025},
  eprint        = {2504.21719},
  archivePrefix = {arXiv}
}

@MISC{jakob2022mitsuba3,
  author       = {Jakob, Wenzel and Speierer, S{\'e}bastien and Roussel, Nicolas and
                  Nimier-David, Merlin and Vicini, Delio and Zeltner, Tizian and
                  Nicolet, Baptiste and Crespo, Miguel and Leroy, Vincent and Zhang, Ziyi},
  title        = {{Mitsuba 3} Renderer},
  howpublished = {Software},
  year         = {2022},
  note         = {Version 3.8.0}
}

@INPROCEEDINGS{loshchilov2019decoupled,
  author    = {Loshchilov, Ilya and Hutter, Frank},
  title     = {Decoupled Weight Decay Regularization},
  booktitle = {Proc. Int. Conf. Learn. Represent. (ICLR)},
  year      = {2019}
}

@ARTICLE{jakob2022drjit,
  author  = {Jakob, Wenzel and Speierer, S{\'e}bastien and Roussel, Nicolas and Vicini, Delio},
  title   = {{Dr.Jit}: A Just-In-Time Compiler for Differentiable Rendering},
  journal = {ACM Trans. Graph.},
  volume  = {41},
  number  = {4},
  pages   = {124:1--124:19},
  month   = jul,
  year    = {2022},
  doi     = {10.1145/3528223.3530099}
}

@INPROCEEDINGS{zhang2024wisegrt,
  author    = {Zhang, Lihao and others},
  title     = {{WiSegRT}: Dataset for Site-Specific Indoor Radio Propagation Modeling with {3D} Segmentation and Differentiable Ray-Tracing: (Invited Paper)},
  booktitle = conf_icnc,
  year      = {2024},
  month     = feb,
  pages     = {744--748},
  doi       = {10.1109/ICNC59896.2024.10556262}
}

@INPROCEEDINGS{hehn2025wigatr,
  author    = {Hehn, Thomas and others},
  title     = {Differentiable and Learnable Wireless Simulation with Geometric Transformers},
  booktitle = conf_iclr,
  year      = {2025},
  month     = apr
}

@MISC{luo2023sensiverse,
  author        = {Luo, Jiang and others},
  title         = {Sensiverse: A Dataset for {ISAC} Study},
  howpublished  = {arXiv:2308.13789},
  year          = {2023},
  eprint        = {2308.13789},
  archivePrefix = {arXiv}
}

@MISC{liu2025openpathnet,
  author        = {Liu, Lizhou and Chen, Xiaohui and Zhang, Wenyi},
  title         = {{OpenPathNet}: An Open-Source {RF} Multipath Data Generator for {AI}-Driven Wireless Systems},
  howpublished  = {arXiv:2512.17286},
  year          = {2025},
  eprint        = {2512.17286},
  archivePrefix = {arXiv}
}

@ARTICLE{li2019directional,
  author  = {Li, Zhuyin and others},
  title   = {Directional Antenna Channel Modeling in Urban Area Using Ray Tracing},
  journal = {{ITU} J. ICT Discoveries},
  year    = {2019},
  month   = dec,
  volume  = {2},
  number  = {1}
}

@INPROCEEDINGS{raddwan2024panda5gsim,
  author    = {Raddwan, Basheer Ameen and Al-Baltah, Ibrahim Ahmed and Ghaleb, Mukhtar},
  title     = {Environment-Aware {3D} Mobility Simulation for the {5G} and {6G} Wireless Networks},
  booktitle = conf_iceti,
  year      = {2024},
  pages     = {1--8},
  doi       = {10.1109/ICETI63946.2024.10777167}
}

@MISC{jiang2025cfm,
  author        = {Jiang, Jun and others},
  title         = {Towards Channel Foundation Models ({CFMs}): Motivations, Methodologies and Opportunities},
  howpublished  = {arXiv:2507.13637},
  year          = {2025},
  eprint        = {2507.13637},
  archivePrefix = {arXiv}
}

@INPROCEEDINGS{alikhani2025lwm,
  author    = {Alikhani, Sadjad and Charan, Gouranga and Alkhateeb, Ahmed},
  title     = {{LWM}: A Pre-Trained Wireless Foundation Model for Universal Feature Extraction},
  booktitle = conf_icmlcn,
  year      = {2025},
  month     = may,
  pages     = {1--6},
  doi       = {10.1109/ICMLCN64995.2025.11140266}
}

@ARTICLE{liu2025wifo,
  author  = {Liu, Boxun and others},
  title   = {{WiFo}: Wireless Foundation Model for Channel Prediction},
  journal = sci_j_scis,
  year    = {2025},
  volume  = {68},
  number  = {6},
  pages   = {162302},
  doi     = {10.1007/s11432-025-4349-0}
}

@ARTICLE{guler2025multitask,
  author  = {Guler, Berkay and Geraci, Giovanni and Jafarkhani, Hamid},
  title   = {A Multi-Task Foundation Model for Wireless Channel Representation Using Contrastive and Masked Autoencoder Learning},
  journal = ieee_j_jsac,
  year    = {2026},
  volume  = {44},
  pages   = {4489--4504},
  doi     = {10.1109/JSAC.2026.3677157}
}

@INPROCEEDINGS{jiang2025csiclip,
  author    = {Jiang, Jun and others},
  title     = {A {MIMO} Wireless Channel Foundation Model via {CIR--CSI} Consistency},
  booktitle = conf_icmlcn,
  year      = {2025},
  month     = may,
  pages     = {1--6},
  doi       = {10.1109/ICMLCN64995.2025.11140262}
}

@ARTICLE{yang2025wirelessgpt,
  author  = {Yang, Tingting and others},
  title   = {{WirelessGPT}: A Generative Pre-Trained Multi-Task Learning Framework for Wireless Communication},
  journal = ieee_m_network,
  year    = {2025},
  month   = sep,
  volume  = {39},
  number  = {5},
  pages   = {58--65},
  doi     = {10.1109/MNET.2025.3579496}
}

@ARTICLE{yang2026wirelessgpt,
  author  = {Yang, Tingting and others},
  title   = {{WirelessGPT}: A Generative Foundation Model for Multi-Task Integrated Sensing and Communication},
  journal = ieee_j_jsac,
  year    = {2026},
  volume  = {44},
  pages   = {2259--2273},
  doi     = {10.1109/JSAC.2025.3640156}
}

@ARTICLE{aboulfotouh2025wavesfm,
  author  = {Aboulfotouh, Ahmed and Mohammed, Elsayed and Abou-Zeid, Hatem},
  title   = {{6G WavesFM}: A Foundation Model for Sensing, Communication, and Localization},
  journal = ieee_j_ojcoms,
  year    = {2025},
  volume  = {6},
  pages   = {6792--6807},
  doi     = {10.1109/OJCOMS.2025.3600616}
}

@ARTICLE{mashaal2025iqfm,
  author  = {Mashaal, Omar Ahmad and Abou-Zeid, Hatem},
  title   = {{IQFM}---A Wireless Foundation Model for {I/Q} Streams in {AI}-Native {6G}},
  journal = ieee_j_ojcoms,
  year    = {2026},
  volume  = {7},
  pages   = {1426--1441},
  doi     = {10.1109/OJCOMS.2026.3661435}
}

@ARTICLE{zheng2025musefm,
  author  = {Zheng, Tianyue and others},
  title   = {{MUSE-FM}: Multi-Task Environment-Aware Foundation Model for Wireless Communications},
  journal = ieee_j_wcom,
  year    = {2026},
  volume  = {25},
  pages   = {19791--19806},
  doi     = {10.1109/TWC.2026.3708709}
}

@MISC{chen2026wirelessworldmodel,
  author        = {Chen, Ziqi and others},
  title         = {A Wireless World Model for {AI}-Native {6G} Networks},
  howpublished  = {arXiv:2603.25216},
  year          = {2026},
  eprint        = {2603.25216},
  archivePrefix = {arXiv}
}

@MISC{zou2026telecomworldmodels,
  author        = {Zou, Hang and others},
  title         = {Telecom World Models: Unifying Digital Twins, Foundation Models, and Predictive Planning for {6G}},
  howpublished  = {arXiv:2604.06882},
  year          = {2026},
  eprint        = {2604.06882},
  archivePrefix = {arXiv}
}

@MISC{moon2026channel2world,
  author        = {Moon, Hyung-Joo and Jang, Joonkyu and Kim, Kwang Soon and
                   Kim, Seong-Lyun and Heath, Jr., Robert W. and Chae, Chan-Byoung},
  title         = {{Channel2World}: A Wireless Foundation Model for {RF} Environment Representation},
  howpublished  = {arXiv:2608.17544},
  year          = {2026},
  eprint        = {2608.17544},
  archivePrefix = {arXiv}
}

@UNPUBLISHED{jang2026channels4world,
  author = {Jang, Joonkyu and Moon, Hyung-Joo and Kim, Kwang Soon and
            Kim, Seong-Lyun and Hoydis, Jakob and Heath, Jr., Robert W. and
            Chae, Chan-Byoung},
  title  = {{Channels4World}: 100,000 {RF} Environments for Wireless Foundation Models},
  note   = {accepted to IEEE GLOBECOM 2026 Workshops, also available at arXiv},
  year   = {2026}
}
\endgroup

\end{document}